\documentclass{article}

    \PassOptionsToPackage{numbers, compress}{natbib}

\usepackage[main, final]{neurips_2026_vericode}

\usepackage[utf8]{inputenc} 
\usepackage[T1]{fontenc}    
\usepackage{hyperref}       
\usepackage{url}   
\usepackage{amssymb}
\newcommand{\pmark}{\textcolor[HTML]{E8912D}{$\blacktriangle$}}
\usepackage{booktabs}       
\usepackage{amsfonts}       
\usepackage{nicefrac}       
\usepackage{xcolor}         
\usepackage{graphicx}
\usepackage{tabularx}
\usepackage{array}
\usepackage{amsmath}
\usepackage{graphicx}
\usepackage{pifont}
\usepackage{xcolor}
\usepackage{fvextra}

\usepackage{longtable}
\usepackage{booktabs}
\usepackage{array}

\usepackage{booktabs}
\usepackage{longtable}
\usepackage{array}
\usepackage{float}
\usepackage{amsmath}
\usepackage{needspace}

\usepackage{algorithm}
\usepackage{algpseudocode}
\usepackage{wrapfig}
\usepackage{tcolorbox}
\usepackage{caption}
\usepackage{makecell}
\usepackage{xcolor}
\usepackage{caption}
\usepackage{kotex}
\usepackage{pgfplots}
\pgfplotsset{compat=1.18}
\usepgfplotslibrary{polar}
\usepackage{multirow}
\usepackage{amsmath}

\newcommand{\PromptTitle}[1]{%
  \Needspace{5\baselineskip}%
  \par
  \vspace{4pt}%
  \noindent
  \begingroup
  \setlength{\fboxsep}{4pt}%
  \colorbox{black!72}{%
    \parbox{\dimexpr\linewidth-2\fboxsep\relax}{%
      \textcolor{white}{\bfseries #1}%
    }%
  }%
  \endgroup
  \par\nobreak
  \vspace{-5pt}%
}

\DefineVerbatimEnvironment{PromptBlock}{Verbatim}{
  breaklines=true,
  breakanywhere=true,
  breaksymbolleft={},
  breaksymbolright={},
  frame=single,
  framesep=1.5mm,
  rulecolor=\color{black!40},
  fontsize=\small,
  listparameters={
    \setlength{\topsep}{0pt}
    \setlength{\partopsep}{0pt}
    \setlength{\parsep}{0pt}
  }
}

\makeatletter
\newenvironment{breakablealgorithm}
  {%
   \begin{center}
   \refstepcounter{algorithm}%
   \hrule height .8pt depth 0pt \kern 2pt%
   \renewcommand{\caption}[2][\relax]{%
     {\raggedright\textbf{\ALG@name~\thealgorithm} ##2\par}%
     \ifx\relax##1\relax
       \addcontentsline{loa}{algorithm}{\protect\numberline{\thealgorithm}##2}%
     \else
       \addcontentsline{loa}{algorithm}{\protect\numberline{\thealgorithm}##1}%
     \fi
     \kern 2pt\hrule\kern 2pt
   }%
  }
  {%
   \kern 2pt\hrule\relax
   \end{center}
  }
\makeatother

\newcommand{\cmark}{\textcolor[HTML]{2E9E4F}{\ding{51}}} 
\newcommand{\xmark}{\textcolor[HTML]{D64545}{\ding{55}}} 

\DeclareRobustCommand{\globalstar}{%
  \textcolor[HTML]{D62839}{\ensuremath{\bigstar}}%
}

\DeclareRobustCommand{\koreanstar}{%
  \textcolor[HTML]{F06A9B}{\ensuremath{\bigstar}}%
}

\title{
\raisebox{-0.2\height}{
    \includegraphics[height=1.7em]{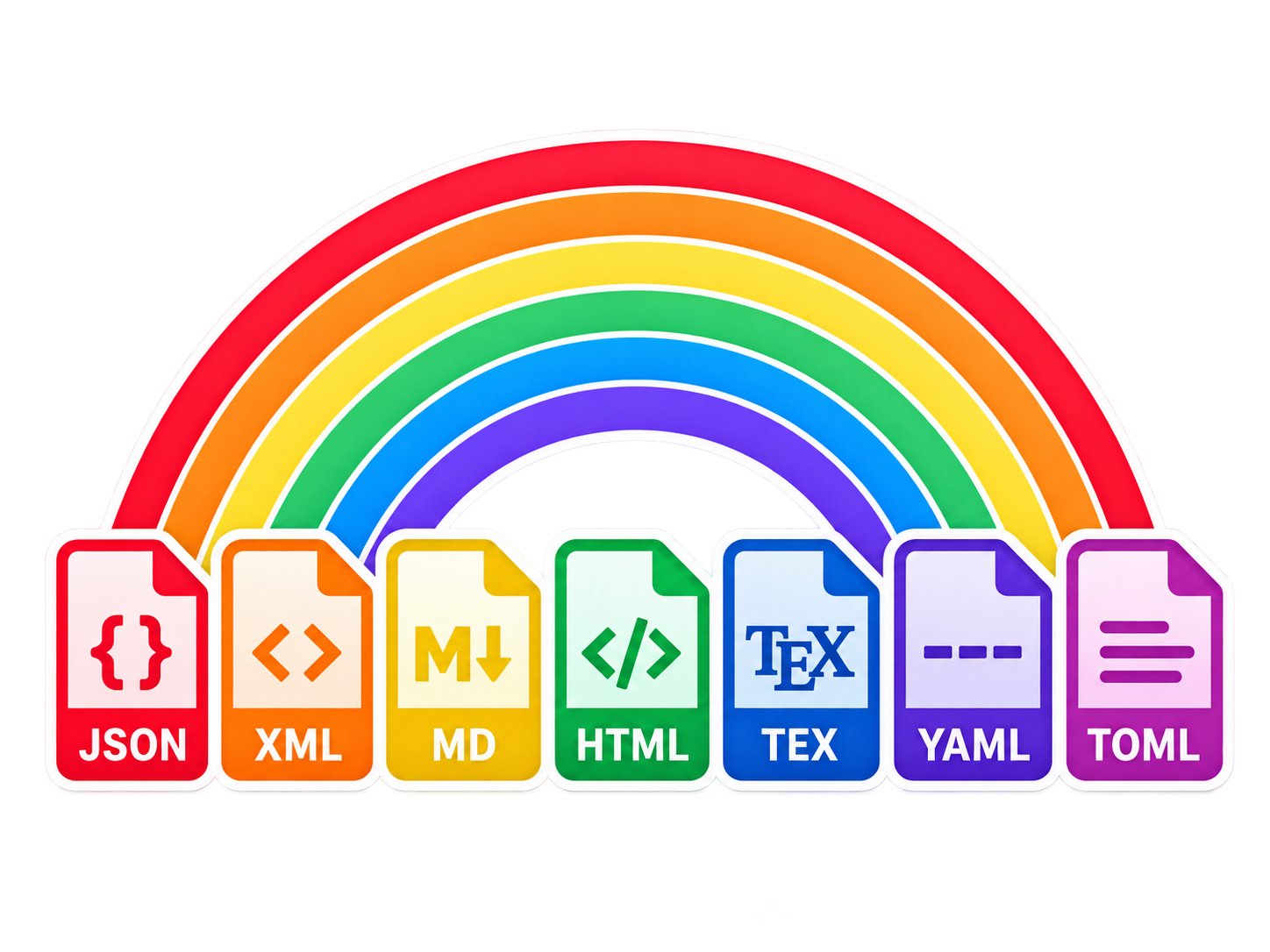}
}\hspace{-2mm}
\textsc{Serial-7}: Deriving, Realizing, and Preserving Machine-Verifiable Outputs Across Representations
}
\author{%
  Youngkyung Seo \quad
  MiHyeon Kim \quad
  Jisoo Lee \quad
  Hwijung Ryu\thanks{Corresponding author: \texttt{hwijung.ryu@kt.com}.} \\
  KT Corporation \\
  \texttt{\{yg.seo, mihyeon.gim, tojs.lee, hwijung.ryu\}@kt.com}
}

\begin{document}
\maketitle

\begin{abstract}
Large language models increasingly generate structured outputs consumed by programs and downstream systems, where semantic correctness alone does not guarantee reliable execution. We introduce SERIAL-7, a Korean--English benchmark built around an executable verification framework for machine-verifiable generation across seven representations. Using 177 validated canonical artifacts, SERIAL-7 separates semantic derivation (\emph{Derive}), constrained artifact generation (\emph{Realize}), and cross-representation preservation (\emph{Preserve}). Its 43 atomic representation constraints are paired with executable checks over format, structure, and task-specific semantics, enabling progressively stricter verification from representation compliance to full artifact validity. Across ten models, REALIZE-L3 HSR ranges from 42.0\% to 93.9\%, while Full Compliance exposes additional artifact-level failures beyond correct final answers and valid representations. These results show that reliable structured generation requires explicit, executable verification of how artifacts are realized and preserved across representations.
\end{abstract}

\begin{figure}[!h]
    \centering
    \includegraphics[width=\linewidth]{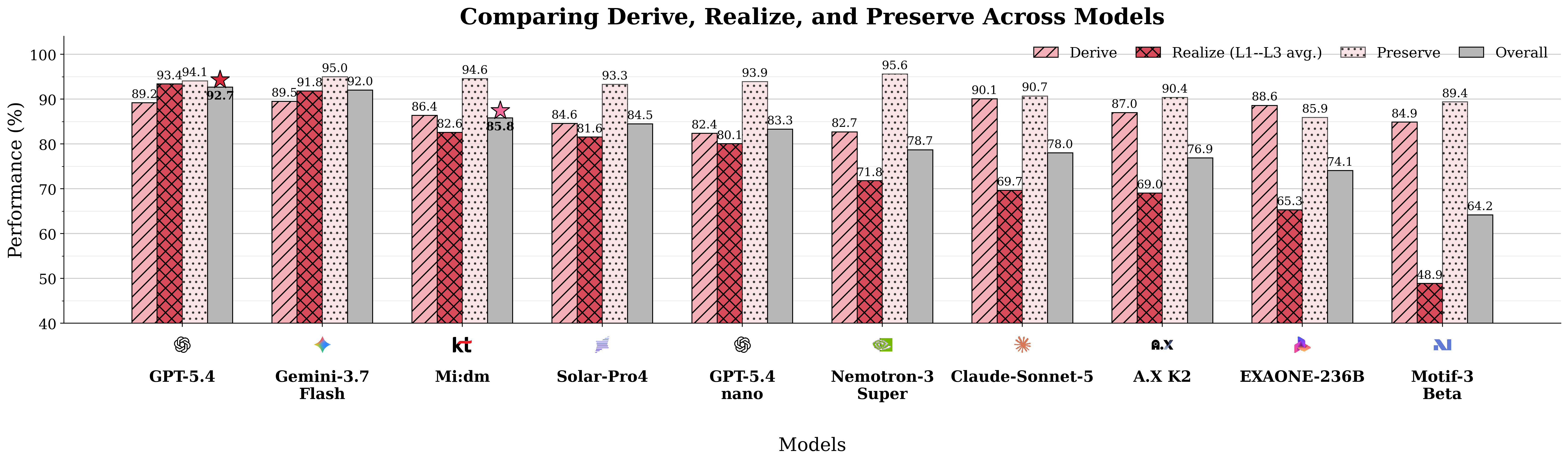}
    \caption{\textbf{Model performance on SERIAL-7.} \globalstar\ and \koreanstar\ denote the best global and Korean-developed models, respectively.}
    \label{fig:modelperformance}
\end{figure}


\section{Introduction}
\label{sec:introduction}

Large language models (LLMs) increasingly produce outputs consumed directly
by programs, verifiers, and downstream systems. Semantic correctness alone is
therefore insufficient: a correct answer may still violate the syntax,
structure, or interface requirements of its target representation
\citep{xia-etal-2024-fofo,wang-etal-2025-verifiable,geng2025jsonschemabench}.
Recent benchmarks cover format following, structured generation,
multi-constraint instruction following, and executable verification
\citep{xia-etal-2024-fofo,yang2026tmlr-structeval,
wang-etal-2025-verifiable,jiang-etal-2024-followbench,
qin-etal-2024-infobench,zhang-etal-2025-cfbench,
qi2025neurips-agentif,moon-etal-2025-mcbench,wang-etal-2023-recode}.
However, they often vary tasks, prompts, schemas, or constraints together,
making it difficult to determine whether a failure originates in semantic
derivation, representation realization, or the validity of the generated
artifact itself. This leaves a basic question unresolved:
\emph{when the semantic target is fixed, how reliably can models realize and
preserve it under machine-verifiable representation contracts, and where do
failures occur?}

We introduce \textsc{SERIAL-7}, a Korean--English benchmark built around an
executable verification framework for studying this question across seven
output representations and 43 atomic representation constraints.
\textsc{SERIAL-7} separates three complementary settings under a fixed semantic
target: \textsc{Derive} measures semantic derivation, \textsc{Realize} evaluates
verified structured generation under explicit representation constraints, and
\textsc{Preserve} evaluates verified cross-representation artifact
transformation. \textsc{Realize} further varies constraint composition through
independently sampled L1--L3 conditions.

For \textsc{Realize} and \textsc{Preserve}, explicit representation contracts
are paired with executable predicates over format, structure, and task-specific
semantic invariants, enabling progressively stricter diagnosis from
representation compliance to primary task success and full artifact-level
validity.

Experiments across ten model configurations reveal that these forms of
reliability are not interchangeable. Realization under explicit representation
contracts varies substantially across models and representations, while
stricter artifact-level verification exposes additional failures beyond
contract compliance and final-answer correctness. \textsc{Preserve} is
generally stronger but exhibits substantial pooled source--target asymmetry.
Representation-, constraint-condition-, and language-dependent differences
further show that reliable structured generation cannot be reduced to answer
accuracy or a single aggregate format-following score. Our main contributions
are:

\begin{itemize}
\item We introduce \textsc{SERIAL-7}, a Korean--English benchmark that holds
the underlying semantic artifact fixed across seven machine-verifiable output
representations.

\item We develop an executable verification framework that maps explicit
representation contracts and 43 atomic constraints to deterministic checks,
enabling failure localization across format, structure, primary task success,
and artifact-level semantic validity.

\item We formulate three complementary evaluation settings---\textsc{Derive},
\textsc{Realize}, and \textsc{Preserve}---that separately probe semantic
derivation, verified structured generation, and verified artifact
transformation, revealing distinct failure patterns under a fixed semantic
target.
\end{itemize}

\begin{figure*}[t]
\centering
\includegraphics[width=\textwidth]{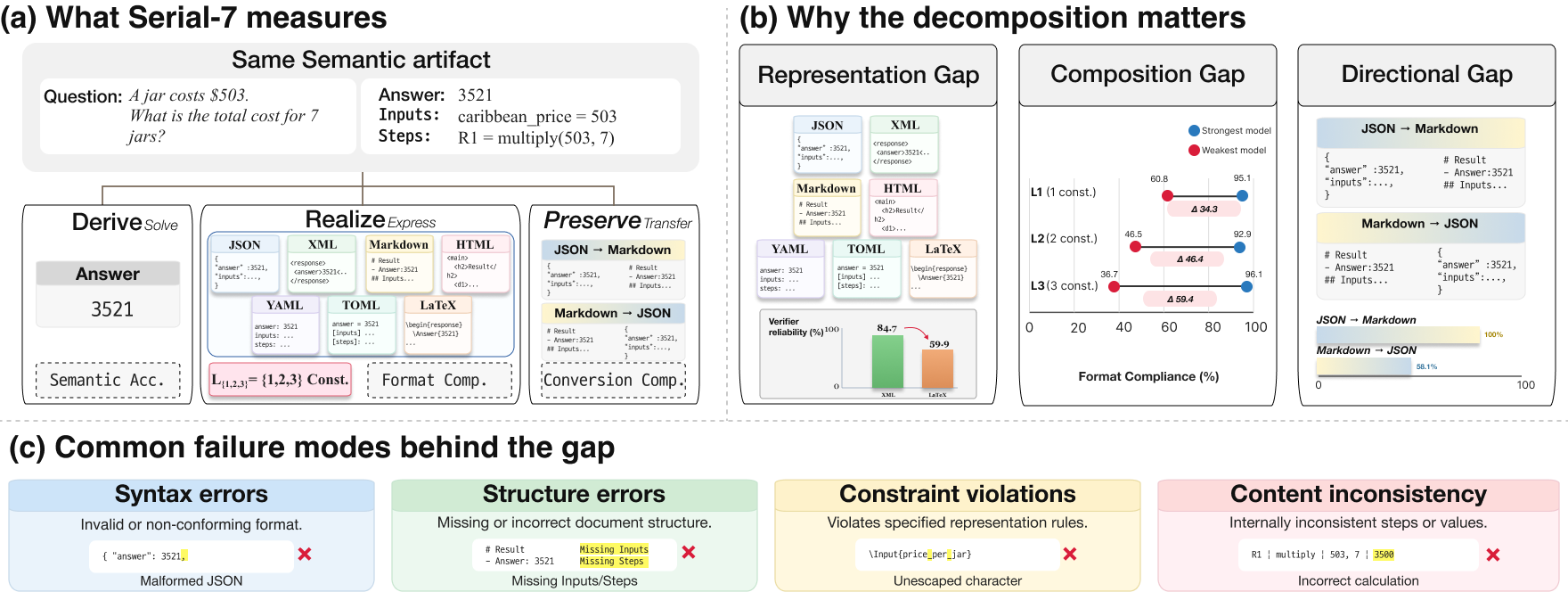}
\caption{
\textbf{Evaluation settings for machine-verifiable generation in
\textsc{SERIAL-7}.}
The same semantic artifact is evaluated through semantic derivation,
verified realization under explicit representation contracts, and
cross-representation artifact preservation. \textsc{Realize} uses a shared
base representation contract, with L1, L2, and L3 additionally activating one,
two, and three independently sampled atomic constraints, respectively.
For structured conditions, executable predicates verify representation,
structure, and task-specific semantic requirements.
}
\label{fig:figure_1}
\end{figure*}
\section{Related Work}

\paragraph{Format and structured-output evaluation.}
Prior work has shown that correct content generation does not necessarily imply
reliable adherence to a required output representation. FoFo evaluates complex
domain-specific formatting instructions \citep{xia-etal-2024-fofo}, while
StructEval broadens structured-output evaluation to multiple textual and
renderable representations and cross-format conversion
\citep{yang2026tmlr-structeval}. ReCode studies a related form of controlled
robustness by applying semantics-preserving perturbations to coding problems
and evaluating the resulting programs through execution
\citep{wang-etal-2023-recode}. These studies establish that representation and
format constraints can substantially affect model behavior. \textsc{Serial-7}
xtends this line of work with a controlled evaluation design that fixes the underlying semantic artifact across paired representations and explicitly separates derivation, constrained realization, and cross-representation preservation.

\paragraph{Verifiable constraint following.}
A complementary line of work focuses on making instruction satisfaction objectively measurable.VFF associates fine-grained format constraints with executable
Python validators, enabling deterministic evaluation of compositional
format requirements \citep{wang-etal-2025-verifiable}.
AgentIF considers more realistic instructions containing multiple
interacting constraints and evaluates compliance at the constraint
level using code-based, model-based, and hybrid evaluators
\citep{qi2025neurips-agentif}. These frameworks make constraint-following failures substantially more observable than aggregate answer accuracy alone. \textsc{Serial-7} builds on this verifiability perspective but asks a different controlled question: how does the required representation itself affect success when the underlying semantic target is held fixed? We therefore pair identical semantic instances across representations and
use executable predicates to separately diagnose format, structural, and
semantic failures under compositional representation constraints.

\begin{table*}[t]
\centering
\small
\setlength{\tabcolsep}{5.0pt}
\caption{
\textbf{Comparison with closely related evaluation frameworks.}
\cmark\ = explicitly evaluated,
\pmark\ = partially or indirectly covered,
and \xmark\ = not explicitly targeted.
\emph{Paired representations} denotes controlled evaluation of the same
underlying semantic target under multiple output representations.
}
\label{tab:related_work}

\resizebox{0.95\textwidth}{!}{
\begin{tabular}{lcccccc}
\toprule
\textbf{Benchmark}
& \textbf{Multiple}
& \textbf{Paired}
& \textbf{Fixed semantic}
& \textbf{Compositional}
& \textbf{Deterministic}
& \textbf{Failure} \\
& \textbf{representations}
& \textbf{representations}
& \textbf{target}
& \textbf{constraints}
& \textbf{verification}
& \textbf{localization} \\
\midrule

FoFo~\citep{xia-etal-2024-fofo}
& \pmark
& \xmark
& \xmark
& \xmark
& \xmark
& \pmark \\

VFF~\citep{wang-etal-2025-verifiable}
& \xmark
& \xmark
& \xmark
& \cmark
& \cmark
& \pmark \\

AgentIF~\citep{qi2025neurips-agentif}
& \xmark
& \xmark
& \xmark
& \cmark
& \pmark
& \cmark \\

StructEval~\citep{yang2026tmlr-structeval}
& \cmark
& \pmark
& \pmark
& \xmark
& \pmark
& \pmark \\

ReCode~\citep{wang-etal-2023-recode}
& \xmark
& \xmark
& \cmark
& \xmark
& \cmark
& \pmark \\

\midrule

\textbf{SERIAL-7 (Ours)}
& \cmark
& \cmark
& \cmark
& \cmark
& \cmark
& \cmark \\

\bottomrule
\end{tabular}
}
\end{table*}

\section{\textsc{SERIAL-7}: Benchmark Construction and Evaluation}

We represent each validated problem as a representation-independent
canonical artifact
\[
z_i=(v_i,I_i,S_i),
\]
where $v_i$ is the final answer, $I_i$ is the ordered sequence of input
identifier--value pairs, and $S_i$ is an optional sequence of computation
steps. Given a representation contract $c$, a deterministic renderer produces
\[
y_{i,c}=R_c(z_i).
\]

For problem $q_i$ and target contract $c_t$, we evaluate
\[
\textsc{Derive}: q_i\rightarrow\hat v_i,\qquad
\textsc{Realize}: (q_i,c_t)\rightarrow\hat y_{i,t},\qquad
\textsc{Preserve}: (y_{i,s},c_t)\rightarrow\hat y_{i,t},
\]
where $y_{i,s}=R_{c_s}(z_i)$.
\textsc{Preserve} additionally requires equality of the required
source-artifact invariants defined in Section~\ref{sec:evaluation_protocol}.
\subsection{Benchmark Construction}
\label{sec:construction}

\paragraph{Source selection and normalization.}
We construct \textsc{Serial-7} from TabMWP~\citep{lu2023tabmwp},
HRM8K~\citep{ko2025hrm8k}, SciBench~\citep{wang2024scibench}, and the STEM
subset of MMLU-Pro~\citep{wang2024mmlupro}. We stratify sampling by source and
initially collect 500 problems. Initial Canonical-IR candidates are generated
with \texttt{gemini-3.1-pro-preview} using a fixed normalization prompt
(Appendix~\ref{app:benchmark_construction}); model outputs are treated only as
candidates and never accepted directly as gold.

\paragraph{Mechanical validation and canonicalization.}
Each candidate undergoes deterministic checks for answer consistency,
grounding, structural validity, graph executability when applicable, and
representation compatibility. Repairable normalization errors are corrected
and fully revalidated; otherwise the candidate is excluded. Items with valid
answers but no verifiable graph are retained as answer-only artifacts.
Appendix~\ref{app:benchmark_construction} gives the full filtering and repair statistics.

\paragraph{Paired Korean--English construction.}
For artifacts with bilingual coverage, Korean and English instances are
derived from the same canonical artifact. Natural-language components are
translated using \texttt{gemini-3.7-flash} and then reviewed by bilingual
annotators, while canonical input identifiers and values, operations, step
structure, and answers remain fixed. The review checks semantic equivalence,
fluency, quantity preservation, computation structure, and source grounding
before an instance is retained for paired evaluation.

\begin{figure*}[t]
    \centering
    \includegraphics[width=0.95\textwidth]{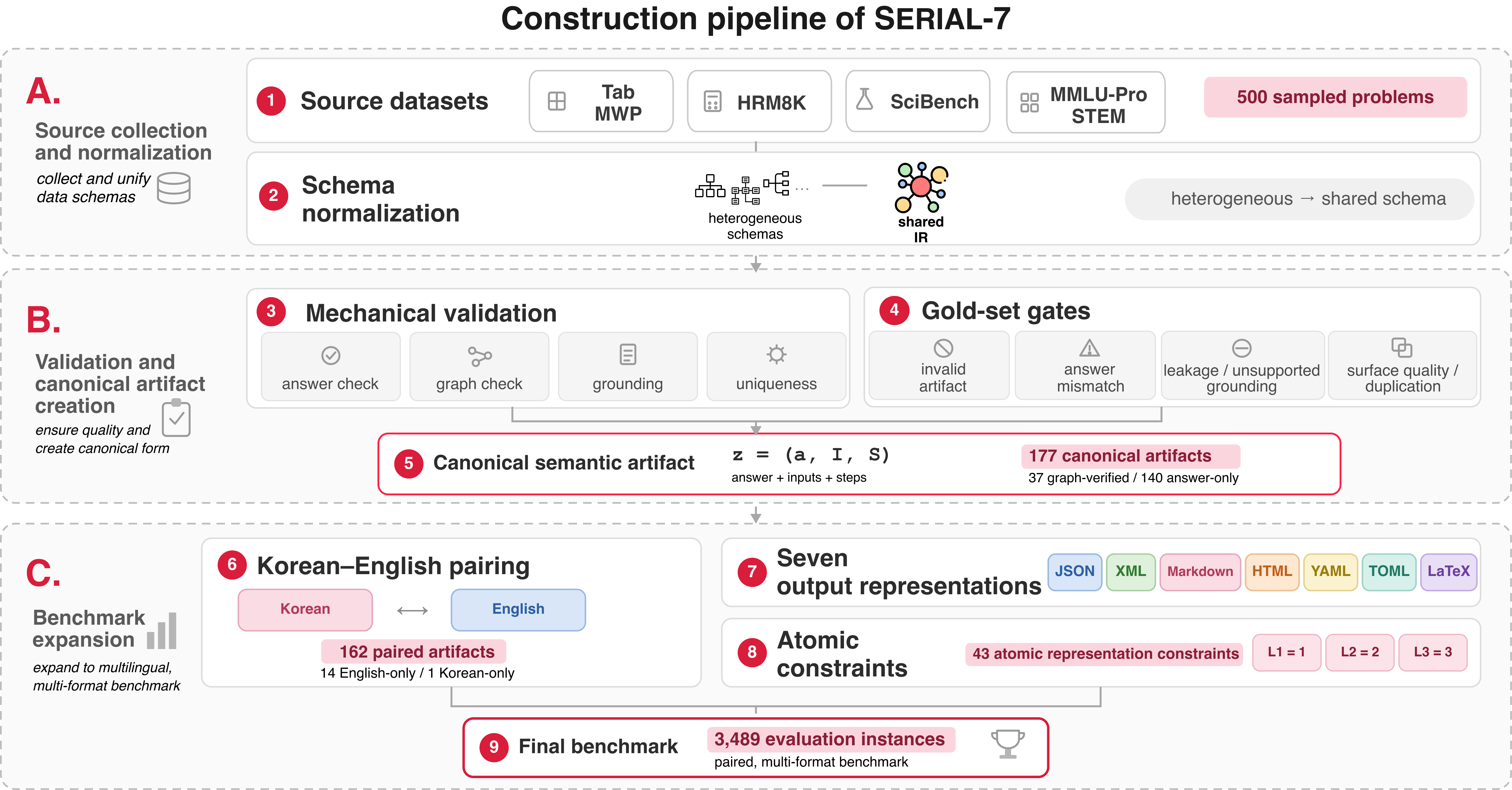}
    \caption{
    \textbf{Construction pipeline of \textsc{Serial-7}.}
    Source problems from four reasoning benchmarks are normalized into
    Canonical-IR candidates, mechanically validated, and, when necessary,
    repaired and revalidated before inclusion. Korean and English evaluation
    instances are constructed from the same validated canonical artifact
    through LLM translation followed by bilingual review.
    }
    \label{fig:construction}
\end{figure*}

\subsection{Representation Contracts}
\label{sec:representation_contracts}

\paragraph{Output representation families.}
Each canonical artifact is deterministically rendered into seven structured,
machine-consumable representations: JSON, XML, Markdown, HTML, \LaTeX{}, YAML,
and TOML. These representations differ in syntax and structural conventions but
encode the same projected semantic content. Representation-specific renderers
control only how the artifact is serialized and are not allowed to introduce
new semantic information.
\paragraph{Atomic representation constraints.}
We define 43 semantics-preserving atomic constraints across seven
representations (Table~\ref{tab:feature_inventory}). Each constraint is
\(a=(C_a,R_a,V_a)\), where \(C_a\) is the instruction clause presented to the
model, \(R_a\) is the deterministic renderer that instantiates the requested
change in the reference artifact, and \(V_a\) is the executable predicate used
to verify the corresponding property in the model output. Constraints may
change representation-specific aspects such as field ordering, container
structure, scalar encoding, or markup, but they do not alter the underlying
semantic content or prescribe a different reasoning decomposition. This
instruction--renderer--verifier coupling makes each requirement explicit,
ensures that the reference realization and evaluation criterion are aligned,
and avoids scoring by surface similarity to a single gold string.

\paragraph{Constraint-count conditions.}
Each target representation has a base contract that specifies its required
machine-verifiable structure. In addition to this base contract,
\textsc{Realize} activates one, two, or three atomic representation
constraints at L1, L2, and L3, respectively. For artifact \(i\),
representation \(r\), and level \(L\), the active atomic-rule set satisfies
\begin{equation}
    |\mathcal{A}_{i,r,L}| = L,
    \qquad
    L \in \{1,2,3\}.
\end{equation}
The rule sets are sampled independently rather than cumulatively, so the levels
should be interpreted as independently sampled constraint-count conditions
rather than a nested or matched-rule difficulty progression. Sampling is
deterministic for each artifact, representation, and level so that all
evaluated models receive the same constraint configuration.

\begin{table*}[t]
\centering
\small
\setlength{\tabcolsep}{5pt}
\renewcommand{\arraystretch}{0.8}
\caption{
\textbf{Atomic representation constraints in \textsc{Serial-7}.}
Constraints modify the representation of a shared canonical artifact without
changing its semantic content.
}
\label{tab:feature_inventory}

\begin{tabularx}{\textwidth}{
    >{\raggedright\arraybackslash}p{1.8cm}
    >{\centering\arraybackslash}p{0.7cm}
    >{\raggedright\arraybackslash}X
}
\toprule
\textbf{Representation} & \textbf{\#} & \textbf{Atomic constraints} \\
\midrule

JSON & 6 &
\texttt{inputs\_as\_map},
\texttt{steps\_as\_map},
\texttt{args\_as\_objects},
\texttt{answer\_object},
\texttt{numeric\_as\_string},
\texttt{key\_order} \\

XML & 5 &
\texttt{args\_as\_attribute},
\texttt{namespace},
\texttt{sibling\_order},
\texttt{value\_as\_attribute},
\texttt{answer\_last} \\

Markdown & 7 &
\texttt{inputs\_as\_table},
\texttt{steps\_as\_table},
\texttt{fence},
\texttt{inline\_code\_ids},
\texttt{heading\_depth},
\texttt{blockquote\_answer},
\texttt{bold\_names} \\

HTML & 6 &
\texttt{table\_for\_inputs},
\texttt{table\_for\_steps},
\texttt{dl\_for\_answer},
\texttt{semantic\_element},
\texttt{strong\_answer},
\texttt{heading\_hierarchy} \\

\LaTeX{} & 6 &
\texttt{tabular\_inputs},
\texttt{enumerate\_steps},
\texttt{siunitx},
\texttt{math\_mode},
\texttt{texttt\_ids},
\texttt{bold\_ids} \\

YAML & 7 &
\texttt{inputs\_as\_map},
\texttt{steps\_as\_map},
\texttt{answer\_object},
\texttt{numeric\_as\_string},
\texttt{key\_order},
\texttt{args\_as\_objects},
\texttt{flow\_args} \\

TOML & 6 &
\texttt{inputs\_as\_table},
\texttt{steps\_as\_table},
\texttt{answer\_table},
\texttt{numeric\_as\_string},
\texttt{key\_order},
\texttt{args\_as\_tables} \\

\midrule
\textbf{Total} & \textbf{43} & \\
\bottomrule
\end{tabularx}
\end{table*}

\subsection{Evaluation Protocol}
\label{sec:evaluation_protocol}
\paragraph{Verification.}

\begin{wrapfigure}[14]{r}{0.47\linewidth}
\vspace{-4.5em}
\centering

\begin{tcolorbox}[
    colback=gray!3,
    colframe=gray!45,
    boxrule=0.5pt,
    arc=1pt,
    left=3pt,
    right=3pt,
    top=4pt,
    bottom=4pt
]
\scriptsize
\textbf{Executable artifact verification}\\[2pt]

\begin{algorithmic}[1]
\Require output $\hat{y}$, format $f$, IR $z$, rules $\mathcal{A}$, view $v$

\State $g \gets
(z.\mathrm{inputs}\neq\emptyset)
\land
(z.\mathrm{steps}\neq\emptyset)$

\State $\mathcal{B}
\gets
\Call{BaseChecks}{f,g,\mathcal{A}}$

\State $\mathbf{b}
\gets
\Call{InitializeFalse}{\mathcal{B}}$

\If{$\mathcal{A}\neq\emptyset$}
    \State $(\hat{z},\mathbf{b})
    \gets
    \Call{FeatureAnalyze}{f,\hat{y},g,\mathcal{A},\mathbf{b}}$
\Else
    \State $(\hat{z},\mathbf{b})
    \gets
    \Call{CarrierAnalyze}{f,\hat{y},g,\mathbf{b}}$
\EndIf

\State $\mathbf{b}[\mathrm{boundary}]
\gets
\Call{CleanBoundary}{f,\hat{y},\mathcal{A}}$

\State $\mathbf{q}
\gets
\left\{
\Call{AtomicCheck}{f,a,\hat{y},\hat{z}}
:
a\in\mathcal{A}
\right\}$

\State $\mathbf{s}
\gets
\Call{StructureChecks}{\hat{z},g}$

\State $\mathbf{m}
\gets
\Call{SemanticChecks}{\hat{z},z,g}$

\State $p_{\mathrm{sem}}
\gets
\Call{ViewSemanticCheck}{v,\hat{z},z,\mathbf{m}}$

\State $\mathrm{FC}
\gets
\Call{AllTrue}{\mathbf{b}\cup\mathbf{q}\cup\mathbf{s}}$

\State $\mathrm{HSR}
\gets
\mathrm{FC}\land p_{\mathrm{sem}}$

\State $\mathrm{Full}
\gets
\mathrm{HSR}\land\Call{AllTrue}{\mathbf{m}}$

\State \Return $(\mathrm{FC},\mathrm{HSR},\mathrm{Full})$
\end{algorithmic}

\end{tcolorbox}

\vspace{-7em}
\end{wrapfigure}

For \textsc{Realize} and \textsc{Preserve}, each structured output is parsed
into the shared Canonical IR and evaluated by executable predicates along
three dimensions: \textbf{format}, covering valid syntax, required containers,
and active representation constraints; \textbf{structure}, covering required
fields and identifier consistency; and \textbf{semantic}, covering the answer,
required input values, and graph consistency when applicable.
For \textsc{Realize}, semantic verification does not require exact
reproduction of the gold reasoning decomposition. For \textsc{Preserve},
\(P_{\mathrm{pres}}(\hat{z},z)\) instead requires source--target equality of
the final answer, ordered input identifiers and values, and, when present,
ordered step identifiers, operators, argument references, and values.
\textsc{Derive} serves as the representation-free semantic reference and does
not activate these mechanical representation checks.
The full carrier-specific verification and rule-aware extraction procedure is
provided in Appendix~\ref{app:executable-verification}.

\paragraph{Metrics.}
\textsc{Derive} is evaluated using Semantic Accuracy. For \textsc{Realize}
and \textsc{Preserve}, we report HSR as a joint contract-and-task success
metric, requiring Format Compliance (FC) together with the view-specific
primary semantic predicate: final-answer correctness for \textsc{Realize}
and \(P_{\mathrm{pres}}\) for \textsc{Preserve}. We additionally report
Full Compliance as the stricter artifact-level criterion, requiring all
applicable executable semantic checks. Accordingly, HSR is not intended to
certify complete artifact semantics on its own; rather, the separation between
FC, HSR, and Full Compliance enables progressively stricter, mechanically
verifiable diagnosis of representation, task, and artifact-level failures.
Formal definitions are provided in Appendix~\ref{app:normalization-scoring}.

\paragraph{Diagnostic measures.}
For diagnostic analysis, we additionally define
\begin{subequations}
\label{eq:diagnostics}
\begin{align}
\Delta_{\mathrm{repr}}^{(m)}
&=
\max_r \mathrm{FC}_{m,r}
-
\min_r \mathrm{FC}_{m,r},
\label{eq:delta-repr}
\\
\Delta_{\mathrm{L1-L3}}^{(m)}
&=
\mathrm{HSR}_{m,L1}
-
\mathrm{HSR}_{m,L3},
\label{eq:delta-l1-l3}
\\
\Delta_{\mathrm{RP}}^{(m)}
&=
\mathrm{HSR}_{m,\mathrm{Preserve}}
-
\mathrm{HSR}_{m,\mathrm{Realize}\text{-}L3},
\label{eq:delta-rp}
\\
\Delta_{\mathrm{pool}}^{(s,t)}
&=
\left|
\overline{\mathrm{HSR}}_{s\rightarrow t}
-
\overline{\mathrm{HSR}}_{t\rightarrow s}
\right|.
\label{eq:delta-pool}
\end{align}
\end{subequations}

Here, \(\mathrm{FC}_{m,r}\) denotes model \(m\)'s mean Format Compliance over
\textsc{Realize} instances with target representation \(r\), aggregated across
L1--L3. \(\mathrm{HSR}_{m,L}\) denotes mean \textsc{Realize} HSR at level
\(L\), and \(\mathrm{HSR}_{m,\mathrm{Preserve}}\) denotes mean HSR over all
observed \textsc{Preserve} instances for model \(m\). For a source--target
transfer \(s\rightarrow t\), \(\overline{\mathrm{HSR}}_{s\rightarrow t}\)
denotes HSR pooled over the observed model-output instances.

\(\Delta_{\mathrm{repr}}\) measures variation across target representations.
\(\Delta_{\mathrm{L1-L3}}\) is the descriptive HSR difference between the
independently sampled L1 and L3 conditions, not a causal effect of adding
constraints. \(\Delta_{\mathrm{RP}}\) is a descriptive cross-condition gap
because \textsc{Realize} and \textsc{Preserve} use different inputs and
evaluation grids. \(\Delta_{\mathrm{pool}}\) measures pooled HSR asymmetry
between observed opposite source--target transfers; because the two transfers
may contain different artifact subsets, it should not be interpreted as an
isolated effect of conversion direction.

\begin{table*}[t]
\centering

\makebox[\textwidth][c]{%

\begin{minipage}[t]{0.42\textwidth}
\vspace{0pt}
\centering

\captionsetup{skip=4pt}
\captionof{table}{\textbf{Canonical gold-set statistics.}}
\label{tab:gold_statistics}

{\footnotesize
\renewcommand{\arraystretch}{0.25}
\setlength{\tabcolsep}{4pt}

\begin{tabularx}{\linewidth}{@{}XX@{}}
\toprule
\multicolumn{1}{c}{\textbf{Source benchmark}} &
\multicolumn{1}{c}{\textbf{Artifact structure}} \\
\midrule

\makecell[l]{
TabMWP: 85\\
HRM8K: 47\\
SciBench: 25\\
MMLU-Pro STEM: 20
}
&
\makecell[l]{
Graph-verified: 37\\
Answer-only: 140
}
\\[2pt]

\midrule

\multicolumn{1}{c}{\textbf{Language availability}} &
\multicolumn{1}{c}{\textbf{Answer characteristics}} \\
\midrule

\makecell[l]{
Korean + English: 162\\
English only: 14\\
Korean only: 1
}
&
\makecell[l]{
Numeric answer: 120\\
Text answer: 57\\
Multiple-choice\\
annotated: 20
}
\\[2pt]

\midrule
\multicolumn{2}{c}{\textbf{Total artifacts: 177}} \\
\bottomrule
\end{tabularx}
}

\end{minipage}%
\hspace{0.015\textwidth}%
\begin{minipage}[t]{0.56\textwidth}
\vspace{0pt}
\centering

\captionsetup{skip=4pt}
\captionof{table}{\textbf{Evaluation instances in \textsc{Serial-7}.}}
\label{tab:instance_statistics}

{\footnotesize
\renewcommand{\arraystretch}{1.00}
\setlength{\tabcolsep}{2pt}

\begin{tabularx}{\linewidth}{
@{}
>{\raggedright\arraybackslash}p{0.12\linewidth}
>{\raggedright\arraybackslash}p{0.27\linewidth}
>{\centering\arraybackslash}p{0.10\linewidth}
>{\raggedleft\arraybackslash}X
@{}
}
\toprule

\textbf{View} &
\textbf{Condition} &
\textbf{\#} &
\textbf{Construction} \\

\midrule

Derive &
Free response &
324 &
162 paired artifacts\newline
$\times$ 2 languages
\\[2pt]

\midrule

Realize &
1--3 active\newline
constraints &
1,470 &
37 graph-verified artifacts\newline
$(33 \times 2 + 4)$ language instances\newline
$\times$ 7 representations $\times$ 3 levels
\\[2pt]

\midrule

Preserve &
Cross-representation\newline
transfer &
1,695 &
$(162 \times 2) + 15$\newline
artifact--language pairs\newline
$\times$ 5 sampled directed transfers
\\[2pt]

\midrule

\multicolumn{2}{l}{\textbf{Total}} &
\textbf{3,489} &
\\

\bottomrule
\end{tabularx}
}

\end{minipage}%
}

\end{table*}

\subsection{Benchmark Statistics}
\label{sec:benchmark_statistics}

\textsc{SERIAL-7} contains 177 validated canonical artifacts and 3,489
evaluation instances; Tables~\ref{tab:gold_statistics} and
\ref{tab:instance_statistics} summarize their construction.


\paragraph{Condition-specific evaluation populations.}
The three conditions use different eligible subsets of the canonical artifact
pool. \textsc{Derive} uses the 162 artifacts with validated Korean--English
pairs, whereas \textsc{Realize} is restricted to the 37 graph-verified
artifacts to maintain a consistent input-, step-, and graph-level semantic
verification scope. Although the verifier supports answer-only carrier checks,
such artifacts do not expose the same step-level semantic structure.
\textsc{Preserve} uses the broader set of 177 validated artifacts when the
required source--target rendering is available. Consequently, scores across
conditions should not be interpreted as paired measurements on an identical
item population.
\section{Experiments}
\label{sec:experiments}
\subsection{Experimental Setup}
\label{sec:experimental_setup}

We evaluate ten model configurations:
GPT-5.4~\citep{openai2026gpt54},
GPT-5.4-nano~\citep{openai2026gpt54nano},
Gemini-3.7-Flash~\citep{google2026gemini37flash},
Claude-Sonnet-5~\citep{anthropic2026claudesonnet5},
Nemotron-3-Super-120B-A12B~\citep{nvidia2026nemotron3super},
Solar-Pro4~\citep{upstage2026solarpro4},
A.X-K2~\citep{baek2026axk2},
K-EXAONE-236B-A23B~\citep{choi2026kexaone},
Motif-3-Beta~\citep{motif2026motif3beta},
and Mi:dm K 3.0 Pro.\footnote{
Mi:dm K 3.0 Pro is a proprietary pre-release model; for preceding public
versions, see Mi:dm 2.0~\citep{shin2026midm20} and Mi:dm K 2.5
Pro~\citep{kt2026midmk25pro}.}
Serving and inference configurations are provided in
Appendix~\ref{app:experimental-configuration}.

We follow the evaluation protocol in
Section~\ref{sec:evaluation_protocol} and the benchmark populations summarized in
Section~\ref{sec:benchmark_statistics}. Because \textsc{Derive},
\textsc{Realize}, and \textsc{Preserve} use different eligible artifact subsets,
cross-condition comparisons are interpreted descriptively rather than as paired
item-level effects. For Korean--English analyses, the canonical artifact is the
statistical unit: matched derived conditions are averaged within each artifact,
and two-sided paired permutation tests are applied with Holm correction across
the ten model-level comparisons.

\begin{figure*}[t]
\centering

\begin{minipage}[t]{0.56\textwidth}
    \vspace{0pt}
    \centering

    \captionsetup{
        font=footnotesize,
        labelfont=bf,
        justification=raggedright,
        singlelinecheck=false,
        skip=5pt
    }

    \captionof{table}{
        \textbf{Performance profiles across \textsc{Serial-7}.}
        \textsc{Derive} reports semantic accuracy;
        \textsc{Realize} and \textsc{Preserve}  operational
joint success (HSR) and strict artifact-level Full Compliance (\%).
        Best and second-best results are shown
        in bold and underlined, respectively.
    }

    \makebox[\linewidth][c]{%
   \resizebox{1.06\linewidth}{!}{%
    \begin{tabular}{@{}lccccc@{}}
    \toprule
    \textbf{Model}
    & \textbf{Derive}
    & \multicolumn{3}{c}{\textbf{Realize}}
    & \textbf{Preserve} \\

    \cmidrule(lr){2-2}
    \cmidrule(lr){3-5}
    \cmidrule(lr){6-6}

    &
    \textit{Acc.}
    & \shortstack{\textbf{R-L1}\\[-1pt]\textit{HSR / Full}}
    & \shortstack{\textbf{R-L2}\\[-1pt]\textit{HSR / Full}}
    & \shortstack{\textbf{R-L3}\\[-1pt]\textit{HSR / Full}}
    & \textit{HSR / Full} \\

    \midrule

    \multicolumn{6}{@{}l}{\textbf{Global Models}} \\

    \quad GPT-5.4
    & 89.2
    & \textbf{94.5} / \textbf{92.8}
    & \textbf{93.3} / \textbf{91.2}
    & \underline{92.4} / \underline{90.1}
    & 94.1 / 93.5 \\

    \quad Gemini-3.7-Flash
    & \underline{89.5}
    & \underline{90.8} / \underline{89.7}
    & \underline{90.8} / \underline{89.9}
    & \textbf{93.9} / \textbf{92.8}
    & \underline{95.0} / \underline{94.4} \\

    \quad GPT-5.4-nano
    & 82.4
    & 85.5 / 81.3
    & 81.6 / 77.2
    & 73.1 / 68.9
    & 93.9 / 92.7 \\

    \quad Nemotron-3-Super-120B-A12B
    & 82.7
    & 78.2 / 73.5
    & 74.5 / 68.1
    & 62.7 / 55.8
    & \textbf{95.6} / \textbf{94.8} \\

    \quad Claude-Sonnet-5
    & \textbf{90.1}
    & 76.3 / 71.2
    & 69.4 / 63.0
    & 63.3 / 56.1
    & 90.7 / 89.9 \\

    \midrule

    \multicolumn{6}{@{}l}{\textbf{Korean-developed Models}} \\

    \quad Mi:dm K 3.0 Pro
    & 86.4
    & \textbf{86.3} / \textbf{83.6}
    & \textbf{82.2} / \textbf{78.7}
    & \underline{79.2} / \underline{74.8}
    & \textbf{95.3} / \textbf{93.8} \\

    \quad Solar-Pro4
    & 84.6
    & \underline{85.5} / \underline{82.7}
    & \underline{79.6} / \underline{76.2}
    & \textbf{79.6} / \textbf{75.3}
    & \underline{93.3} / \underline{92.4} \\

    \quad A.X-K2
    & \underline{87.0}
    & 75.3 / 70.8
    & 69.6 / 64.4
    & 62.2 / 56.8
    & 90.4 / 89.6 \\

    \quad K-EXAONE-236B-A23B
    & \textbf{88.6}
    & 69.2 / 64.7
    & 67.3 / 61.2
    & 59.4 / 52.1
    & 85.9 / 84.8 \\

    \quad Motif-3-Beta
    & 84.9
    & 55.7 / 49.8
    & 49.0 / 42.7
    & 42.0 / 35.1
    & 89.4 / 88.2 \\

    \bottomrule
    \end{tabular}%
    }}

    \label{tab:capability-profile}
\end{minipage}%
\hfill%
\begin{minipage}[t]{0.39\textwidth}
    \vspace{0pt}
    \centering

    \includegraphics[
        width=\linewidth
    ]{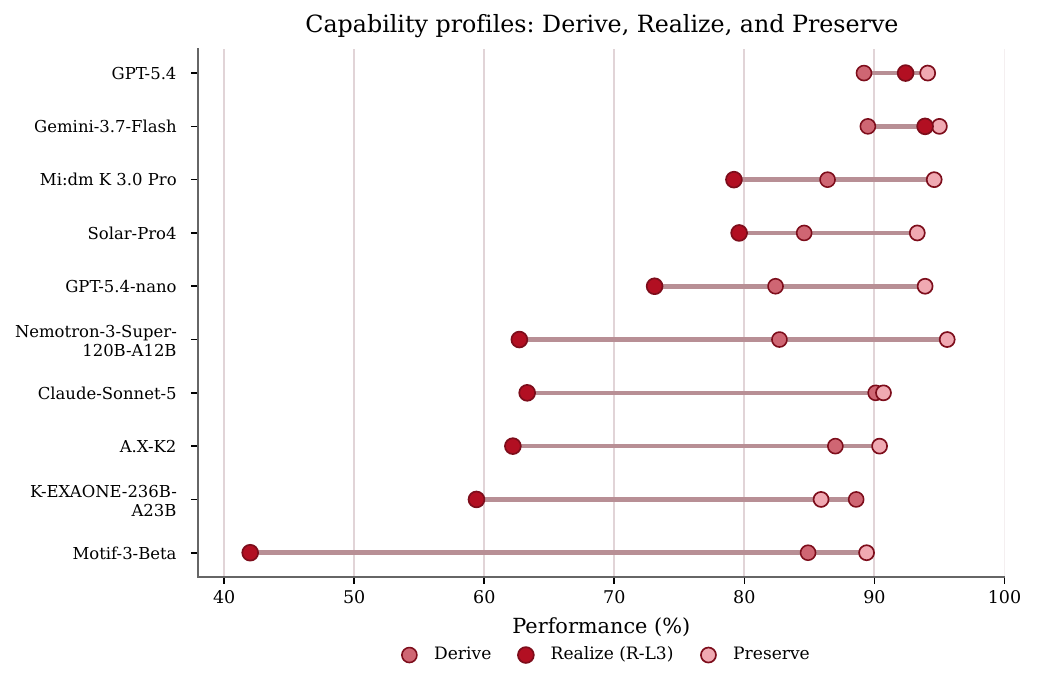}

    \captionsetup{
        font=footnotesize,
        labelfont=bf,
        justification=raggedright,
        singlelinecheck=false,
        skip=5pt
    }

\captionof{figure}{
    \textbf{Derive, Realize, and Preserve performance profiles.}
    \textsc{Derive} reports semantic accuracy; structured conditions
    report HSR. R-L3 denotes the independently sampled
    three-constraint \textsc{Realize} condition.
}
    \label{fig:derive-realize-preserve}
\end{minipage}

\end{figure*}

\subsection{Main Results}
\label{sec:main_results}

Because decoding settings differ across serving configurations
(Appendix~\ref{app:experimental-configuration}), all cross-model comparisons
below are descriptive comparisons of complete model-serving configurations.
Human validation supports evaluator reliability at the aggregate level but is
not used to claim that small score differences constitute resolved rankings.

\begin{figure*}[t]
\centering

\begin{minipage}[t]{0.64\textwidth}
    \centering
    \includegraphics[width=\linewidth]
    {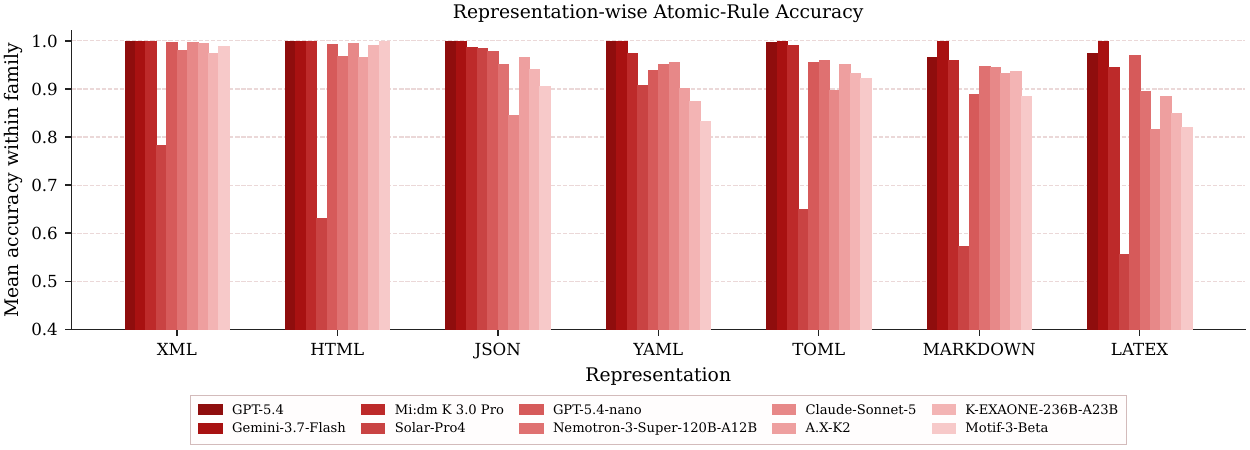}

    \vspace{2pt}
    \small
    \textbf{(a)} Representation-wise atomic-rule accuracy
\end{minipage}
\hfill
\begin{minipage}[t]{0.34\textwidth}
    \centering
    \includegraphics[width=\linewidth]
    {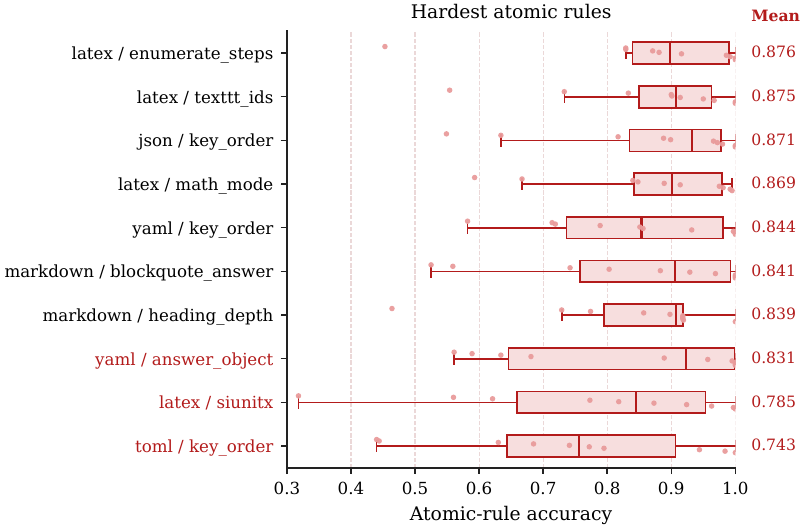}

    \vspace{2pt}
    \small
    \textbf{(b)} Hardest atomic rules across models
\end{minipage}

\caption{
\textbf{Executable atomic-rule diagnostics for constrained realization.}
\textbf{(a)} Representation-wise atomic-rule accuracy, computed only over
instances in which the corresponding rule is active.
\textbf{(b)} Cross-configuration accuracy distributions for the ten most
difficult atomic constraints, ranked by mean accuracy across configurations.
}

\label{fig:atomic-failures}
\end{figure*}

\begin{figure*}[t]
\centering

\begin{minipage}[t]{0.47\textwidth}
    \vspace{0pt}
    \centering

    \includegraphics[
        width=\linewidth
    ]{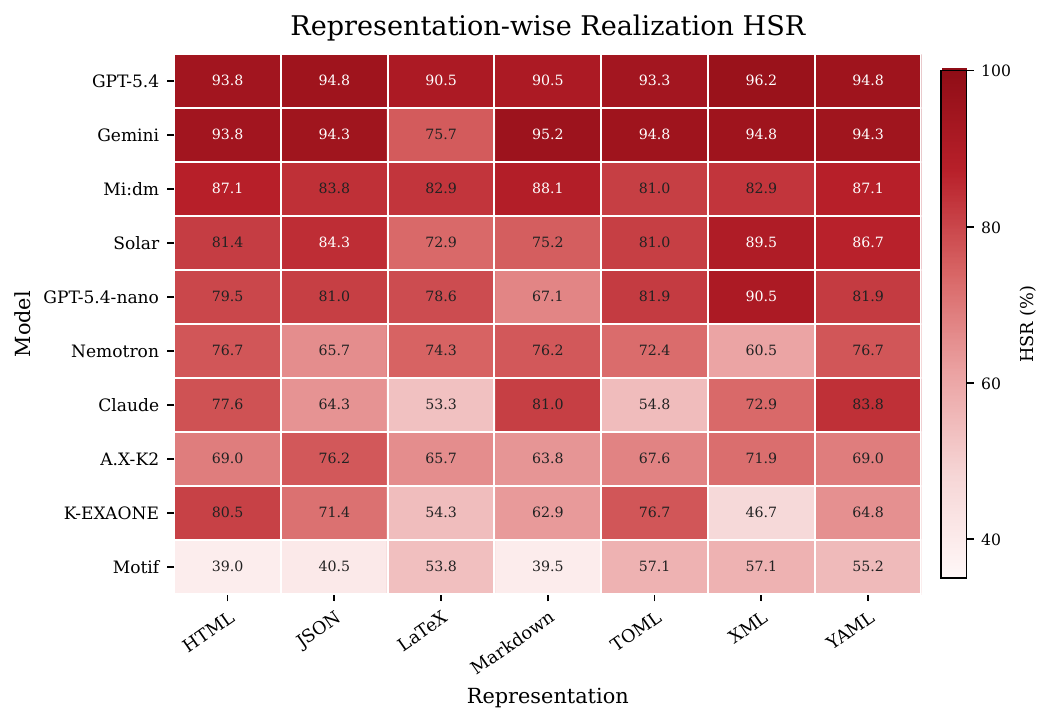}

    \vspace{2pt}
    \small
    \textbf{(a)} Representation-wise Realization HSR
\end{minipage}%
\hfill%
\begin{minipage}[t]{0.47\textwidth}
    \vspace{0pt}
    \centering

    \includegraphics[
        width=\linewidth
    ]{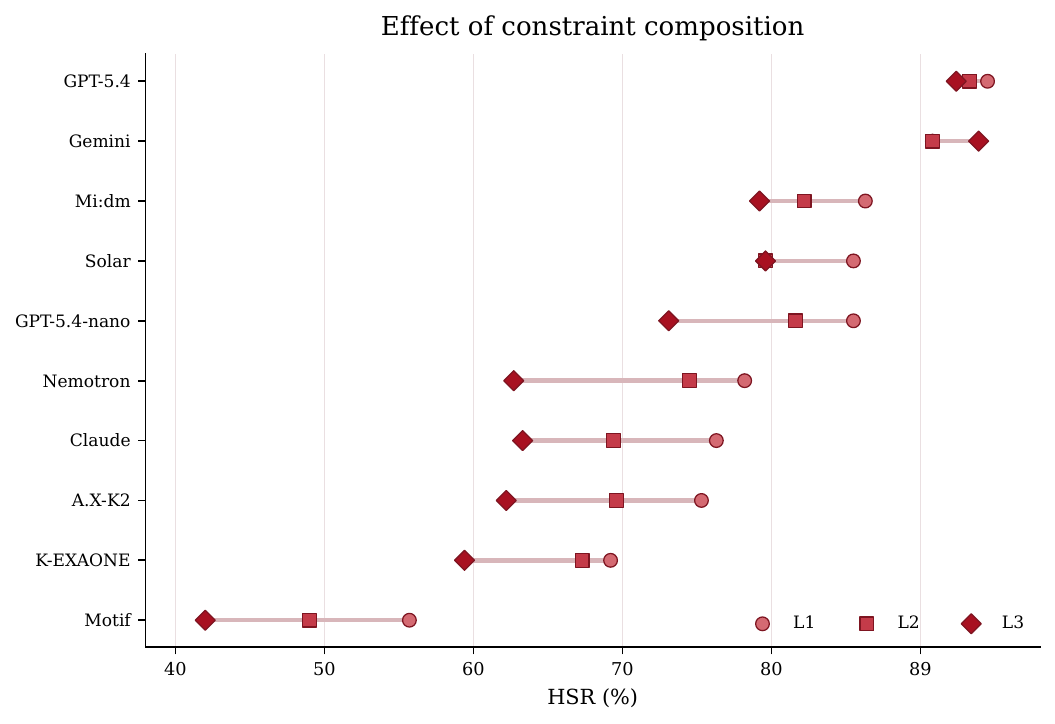}

    \vspace{2pt}
    \small
    \textbf{(b)} HSR across Constraint-Count Conditions
\end{minipage}

\caption{
\textbf{Verified realization under representation constraints.}
\textbf{(a)} Representation-wise \textsc{Realize} HSR across model-serving
configurations and target representations.
\textbf{(b)} HSR under independently sampled one-, two-, and
three-constraint conditions. Rule sets are sampled independently across
levels rather than cumulatively.
}
\label{fig:realize-analysis}

\end{figure*}

\paragraph{Derive, Realize, and Preserve Expose Distinct Reliability Profiles}
\label{sec:capability-separation}

Table~\ref{tab:capability-profile} shows distinct reliability profiles across
the three conditions. Representation-free \textsc{Derive} accuracy occupies a
relatively narrow range (82.4--90.1\%), whereas \textsc{Realize}-L3 HSR spans
42.0--93.9\%. Full Compliance provides a stricter artifact-level criterion,
separating primary task success from satisfaction of all applicable semantic
checks. Thus, similar semantic performance does not imply similar
machine-verifiable structured generation.

Observed rankings also change across conditions. Claude-Sonnet-5 has the
highest observed \textsc{Derive} score (90.1\%) but reaches 63.3\% HSR at
\textsc{Realize}-L3, whereas Nemotron-3-Super reaches 95.6\% under \textsc{Preserve} despite
62.7\% at L3. Similar cross-condition differences are observed among the
Korean-developed models.

These cross-condition differences are descriptive rather than paired effects:
\textsc{Realize} and \textsc{Preserve} differ in artifact population, input
form, and representation-constraint setting. They therefore characterize
distinct verified-generation and verified-transformation profiles rather than
the causal effect of removing semantic derivation.

\paragraph{Where Do Representation-Contract Failures Concentrate?}
\label{sec:atomic-failures}

We compute each atomic predicate's conditional pass rate only over instances
in which it is active; per-rule activation counts are reported in
Appendix~\ref{app:atomic-feature-performance}. Among the observed activations,
the lowest mean pass rates occur for
\texttt{toml.key\_order} (0.743),
\texttt{latex.siunitx} (0.785),
\texttt{yaml.answer\_object} (0.831), and
\texttt{markdown.heading\_depth} (0.839).
Because each requirement has a corresponding executable predicate, these
conditional results identify representation constraints that are frequently
violated within their sampled support. For the similarly defined
\texttt{key\_order} rule, accuracy is 0.871 in JSON, 0.844 in YAML, and
0.743 in TOML, indicating substantial carrier-specific variation.

Failure mechanisms also differ by representation. Some low-scoring rules are
associated with malformed or unrecoverable outputs, whereas others are violated
within otherwise valid documents. Similar conditional pass rates can therefore
reflect either failure to produce a mechanically recoverable artifact or
incomplete execution of an otherwise well-formed representation contract.
Per-rule support and detailed failure analysis are provided in
Appendix~\ref{app:atomic-feature-performance} and
Appendix~\ref{app:atomic-results}, respectively.

\paragraph{Realizing under Representation Constraints}
\label{sec:realize-results}

Figure~\ref{fig:realize-analysis}(a) shows strong
configuration--representation interactions in verified realization.
GPT-5.4 shows relatively small variation across carriers,
Gemini-3.7-Flash shows lower HSR on \LaTeX{}, and Solar-Pro4
achieves higher HSR on XML than on several other representations.
The likelihood that a generated artifact satisfies both its target contract
and primary semantic requirement is therefore configuration- and
representation-dependent rather than globally ordered.
\begin{wrapfigure}[14]{r}{0.40\columnwidth}
    \vspace{-1em}
    \centering

    \includegraphics[
        width=0.7\linewidth
    ]{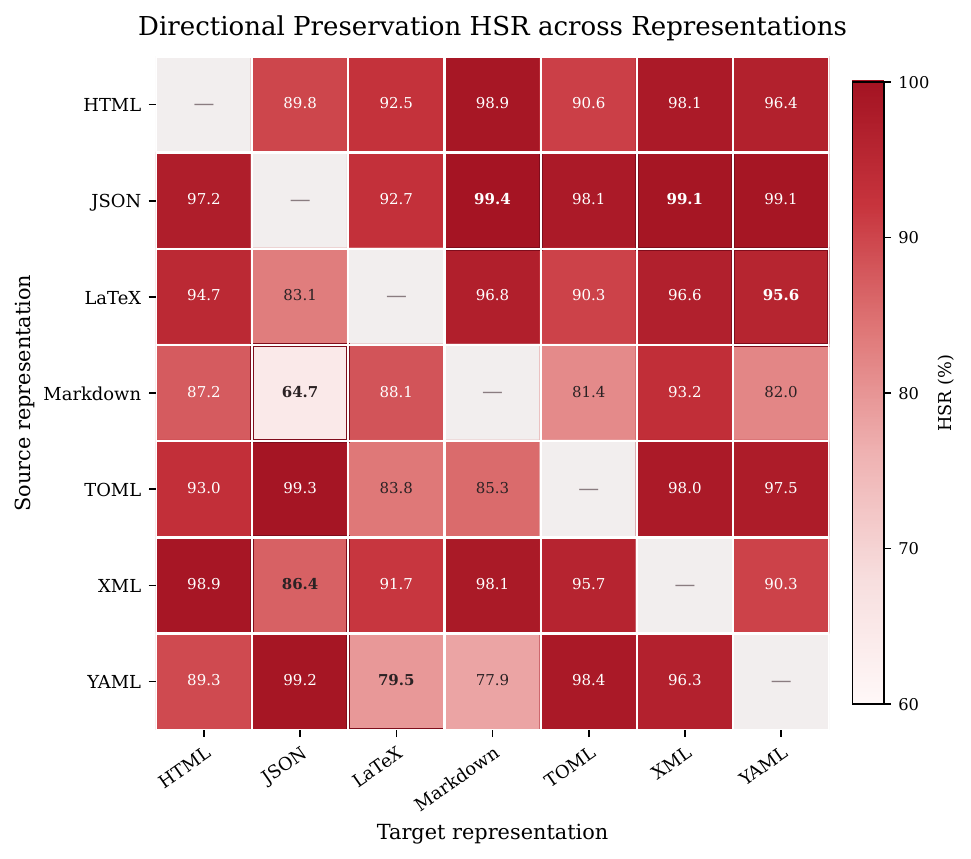}

    \captionsetup{
        font=footnotesize,
        labelfont=bf,
        justification=raggedright,
        singlelinecheck=false,
        skip=3pt
    }

    \caption{
        \textbf{Pooled source--target Preservation HSR.}
        Rows denote source representations and columns denote targets.
        Each cell pools only artifacts for which the corresponding transfer
        was sampled; opposite transfers therefore need not contain identical
        artifact sets.
    }

    \label{fig:preserve-analysis}

    \vspace{-4em}
\end{wrapfigure}
Figure~\ref{fig:realize-analysis}(b) shows that nine of the ten configurations
have lower HSR at L3 than at L1. GPT-5.4 differs by only about two percentage
points, whereas Nemotron-3-Super, Motif-3-Beta, A.X-K2, Claude-Sonnet-5, and
GPT-5.4-nano differ by more than 12 points; Gemini-3.7-Flash is the sole
exception. Because L1--L3 rule sets are sampled independently rather than
cumulatively, these differences do not define a nested or matched-rule
difficulty progression. They describe verified realization under independently
sampled one-, two-, and three-constraint conditions.

\paragraph{Preserving across Representations}
\label{sec:preserve-results}

Model-level \textsc{Preserve} HSR ranges from 85.9\% to 95.6\%, while
\textsc{Realize}-L3 HSR ranges from 42.0\% to 93.9\%. These ranges are not
directly comparable because the conditions differ in artifact population,
input form, and representation-constraint setting. \textsc{Preserve} further
requires source--target preservation of the required artifact content.

Figure~\ref{fig:preserve-analysis} reveals substantial pooled source--target
asymmetry in verified artifact transformation. For example,
JSON$\rightarrow$Markdown reaches 0.994 HSR, whereas
Markdown$\rightarrow$JSON reaches 0.647. Because only five directed transfers are sampled per artifact, opposite
transfers are not necessarily evaluated on identical artifact subsets. Accordingly, Figure~\ref{fig:preserve-analysis} characterizes pooled source--target
asymmetry over the benchmark's sampled transfer grid rather than a
matched-pair estimate of directionality. Similar pooled asymmetries occur for
YAML--\LaTeX{} and XML--JSON. These patterns are consistent with preservation
performance varying with both source recovery and target-contract realization.

\paragraph{Language Effects}
\label{sec:language-results}

\begin{figure}[t]
\centering
\includegraphics[width=\linewidth]
{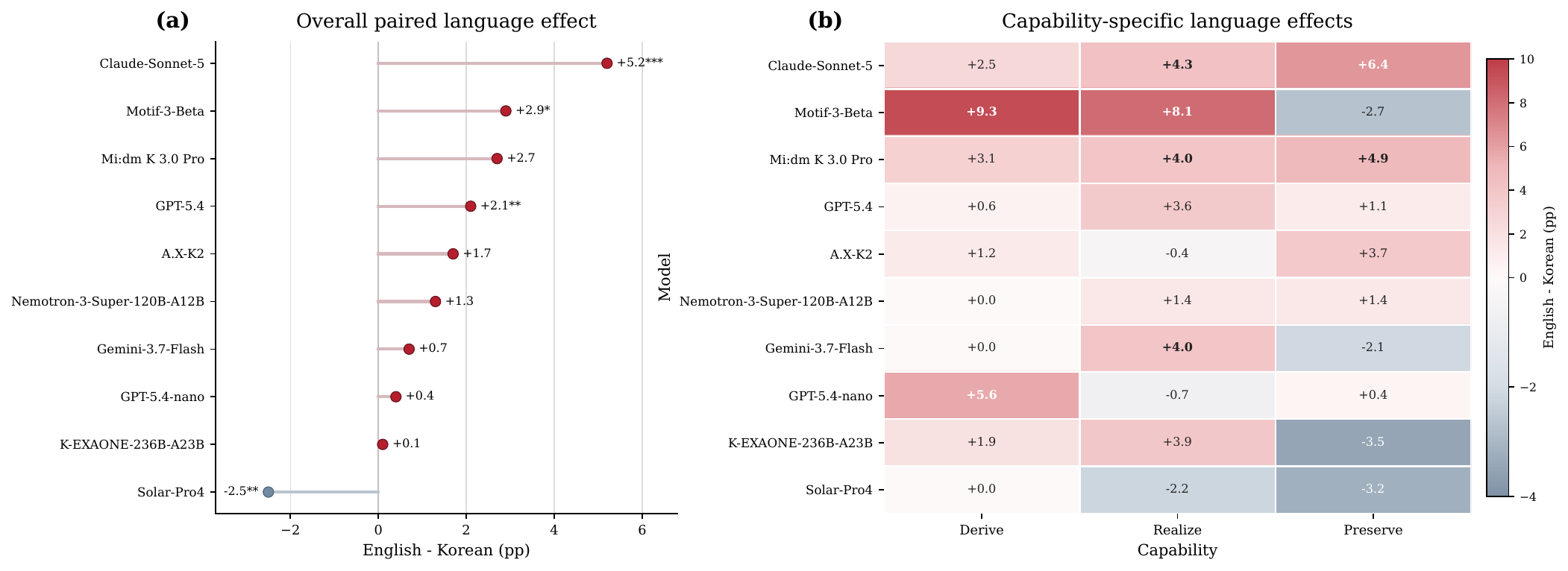}
\caption{
\textbf{Paired English--Korean performance differences.}
\textbf{(a)} Aggregate configuration-level gaps; stars denote significance
under two-sided paired permutation tests using canonical artifacts as the
statistical unit, with Holm correction across models.
\textbf{(b)} Condition-specific gaps across \textsc{Derive},
\textsc{Realize}, and \textsc{Preserve}, reported descriptively.
Positive values favor English.
}
\label{fig:language-gap}
\end{figure}

Because Korean and English instances share the same Canonical IR,
\textsc{Serial-7} enables paired language-conditioned comparisons under an
aligned semantic target. Figure~\ref{fig:language-gap}(a) shows generally
modest aggregate differences, although their magnitude varies across
configurations. Claude-Sonnet-5 shows the largest English advantage
($+5.2$~pp), while Solar-Pro4 shows the clearest Korean advantage
($-2.5$~pp). Significance is assessed from canonical-artifact-level paired
differences rather than treating derived prompts as independent observations.

Figure~\ref{fig:language-gap}(b) further shows that aggregate differences can
mask condition-specific reversals. Gemini-3.7-Flash and
K-EXAONE-236B-A23B favor English under \textsc{Realize} but Korean under
\textsc{Preserve}, while Motif-3-Beta favors English in \textsc{Derive} and
\textsc{Realize} but Korean in \textsc{Preserve}. These results characterize distinct verified-generation and verified-transformation profiles; as specified in Section 3.4, the conditions differ in eligible artifact populations and are not interpreted as causal interventions.

\section{Conclusion}

We introduced \textsc {SERIAL-7} to examine whether strong semantic accuracy translates
into reliable machine-verifiable artifact generation across seven
representations. By separating \textsc{Derive}, \textsc{Realize}, and
\textsc{Preserve}, and pairing representation contracts with executable checks
over format, structure, and task-specific semantic invariants, \textsc{SERIAL-7}
supports progressively stricter evaluation from representation compliance to
full artifact-level validity. Across ten model configurations,
\textsc{Realize}-L3 HSR ranges from 42.0\% to 93.9\%, while Full Compliance
reveals additional mechanically detectable failures beyond contract compliance
and final-answer correctness; \textsc{Preserve} further exhibits substantial
source--target asymmetries. These results show that reliable structured
generation cannot be reduced to answer accuracy alone, but requires explicit
verification of how artifacts are realized and preserved across representations.
\textsc{Serial-7} therefore provides a controlled, executable framework for identifying where machine-verifiable generation succeeds or fails across representation contracts.


\bibliographystyle{plainnat}
\bibliography{neurips_2026}

\appendix
\raggedbottom
\section{Extended Representation Results}
\label{app:extended-results}

This section provides detailed diagnostics underlying the aggregate results in
Section~\ref{sec:main_results}. The main \textsc{Preserve} results use HSR as
the end-to-end success metric, while the analyses below use Format Compliance
to isolate source--target representation variation from semantic correctness.
For \textsc{Realize}, we report conditional atomic-rule pass rates together
with their observed activation support.

\subsection{Full Source--Target Matrices}
\label{app:full-conversion-matrices}

\begin{figure*}[t]
    \centering
    \includegraphics[width=\textwidth]
    {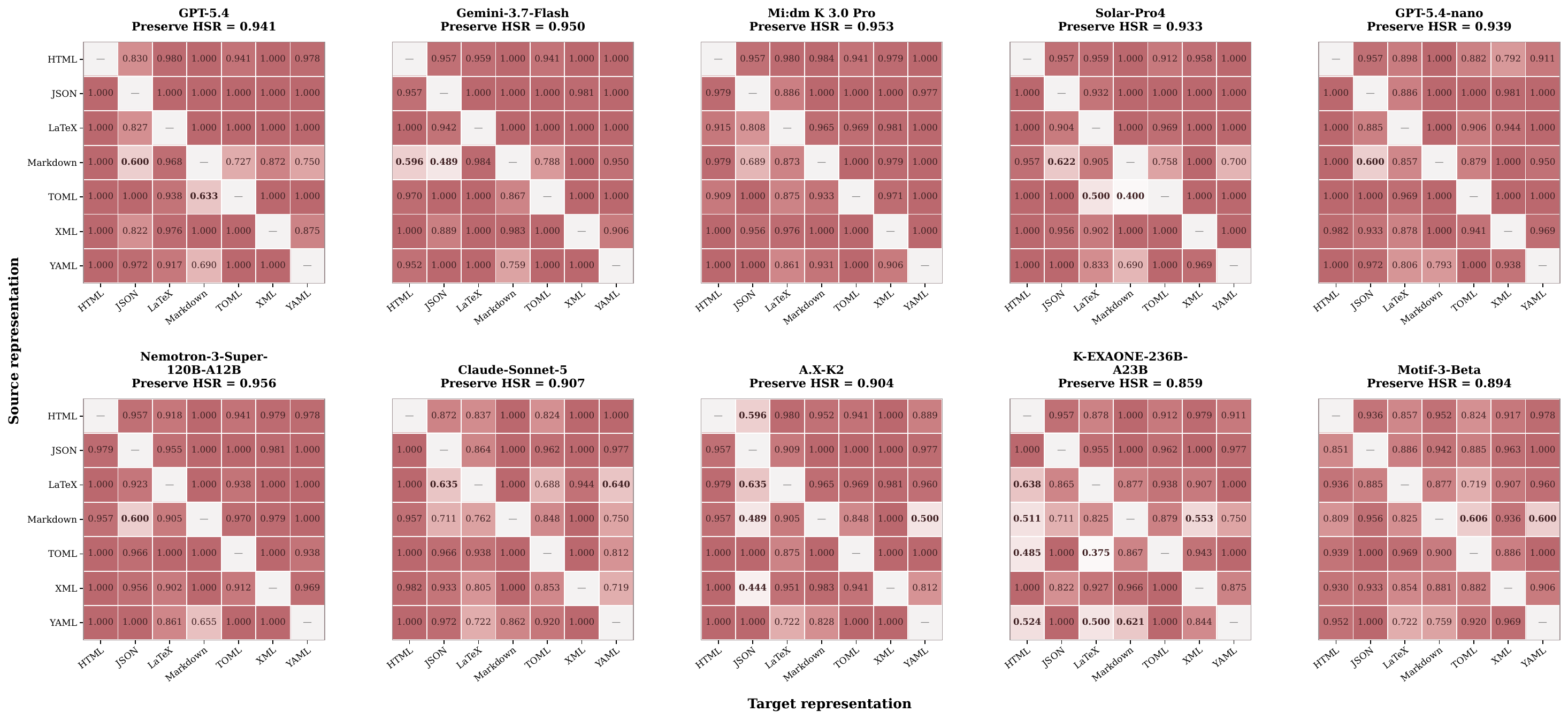}
    \caption{
    \textbf{Per-model source--target preservation matrices.}
    Rows indicate source representations and columns indicate target
    representations. Cells report target-side Format Compliance over the
    sampled instances available for each source--target pair.
    }
    \label{fig:full-conversion-matrices}
\end{figure*}

Figure~\ref{fig:full-conversion-matrices} reports the complete preservation
matrix for each evaluated model. Rows denote the source representation and
columns denote the required target representation; each cell reports
target-side Format Compliance. Diagonal cells are omitted because
\textsc{Preserve} evaluates transfer between different representations.

These format-only matrices complement the pooled HSR matrix in the main paper
by isolating adherence to the target representation contract. Preservation
errors are not uniformly distributed across observed source--target pairs, and
several low-scoring pairs recur across models, although their magnitude varies
by model.

The conversion grid samples five directed representation pairs for each
language-specific artifact rather than all 42 possible directions. The same
sampled grid is used for every evaluated model. Table~\ref{tab:preserve-support}
reports the resulting support once per directed pair, where each count denotes
the number of artifact--language instances per model. The 42 off-diagonal
counts sum to 1,695.

\begin{table}[t]
\centering
\small
\caption{
\textbf{Support for sampled \textsc{Preserve} transfers.}
Each cell reports the number of artifact--language instances per model for
which the corresponding directed transfer is evaluated. The same transfer
grid is shared across all ten model configurations.
}
\resizebox{\linewidth}{!}{
\begin{tabular}{lrrrrrrr}
\toprule
\textbf{Source}
& \textbf{JSON}
& \textbf{XML}
& \textbf{Markdown}
& \textbf{HTML}
& \textbf{\LaTeX{}}
& \textbf{YAML}
& \textbf{TOML} \\
\midrule

JSON
& -- & 34 & 36 & 36 & 40 & 38 & 40 \\

XML
& 40 & -- & 45 & 39 & 45 & 36 & 45 \\

Markdown
& 38 & 42 & -- & 42 & 44 & 41 & 44 \\

HTML
& 43 & 44 & 40 & -- & 36 & 36 & 41 \\

\LaTeX{}
& 43 & 41 & 42 & 42 & -- & 44 & 38 \\

YAML
& 44 & 38 & 39 & 39 & 36 & -- & 38 \\

TOML
& 41 & 38 & 38 & 44 & 44 & 41 & -- \\

\bottomrule
\end{tabular}
}
\label{tab:preserve-support}
\end{table}

\subsection{Pooled Source--Target Asymmetry}
\label{app:directional-asymmetry}

To summarize format-level differences between opposite source--target
transfers, we define the pooled pairwise Format Compliance gap for model $m$
and an unordered representation pair $\{s,t\}$ as
\[
\delta^{FC}_m(s,t)
=
\left|
FC_m(s\rightarrow t)
-
FC_m(t\rightarrow s)
\right|.
\]

We summarize model-level pooled asymmetry by averaging over the set
$\mathcal{U}_m$ of representation pairs for which both transfer directions are
observed:
\[
\Delta_{\mathrm{pool},FC}^{(m)}
=
\frac{1}{|\mathcal{U}_m|}
\sum_{\{s,t\}\in\mathcal{U}_m}
\delta^{FC}_m(s,t).
\]

Because opposite transfers need not contain identical artifact subsets,
$\Delta^{(m)}_{\mathrm{pool},FC}$ is defined as a pooled source--target
asymmetry measure over the sampled transfer grid. It is intended to
characterize benchmark-level differences between opposite transfers rather
than matched item-level differences.
The corresponding support for each transfer is reported in
Table~\ref{tab:preserve-support}. This Format-Compliance measure also differs
from the HSR-based pooled gap in the main analysis by excluding final-answer
correctness.

\begin{figure}[t]
    \centering
    \includegraphics[width=\linewidth]
    {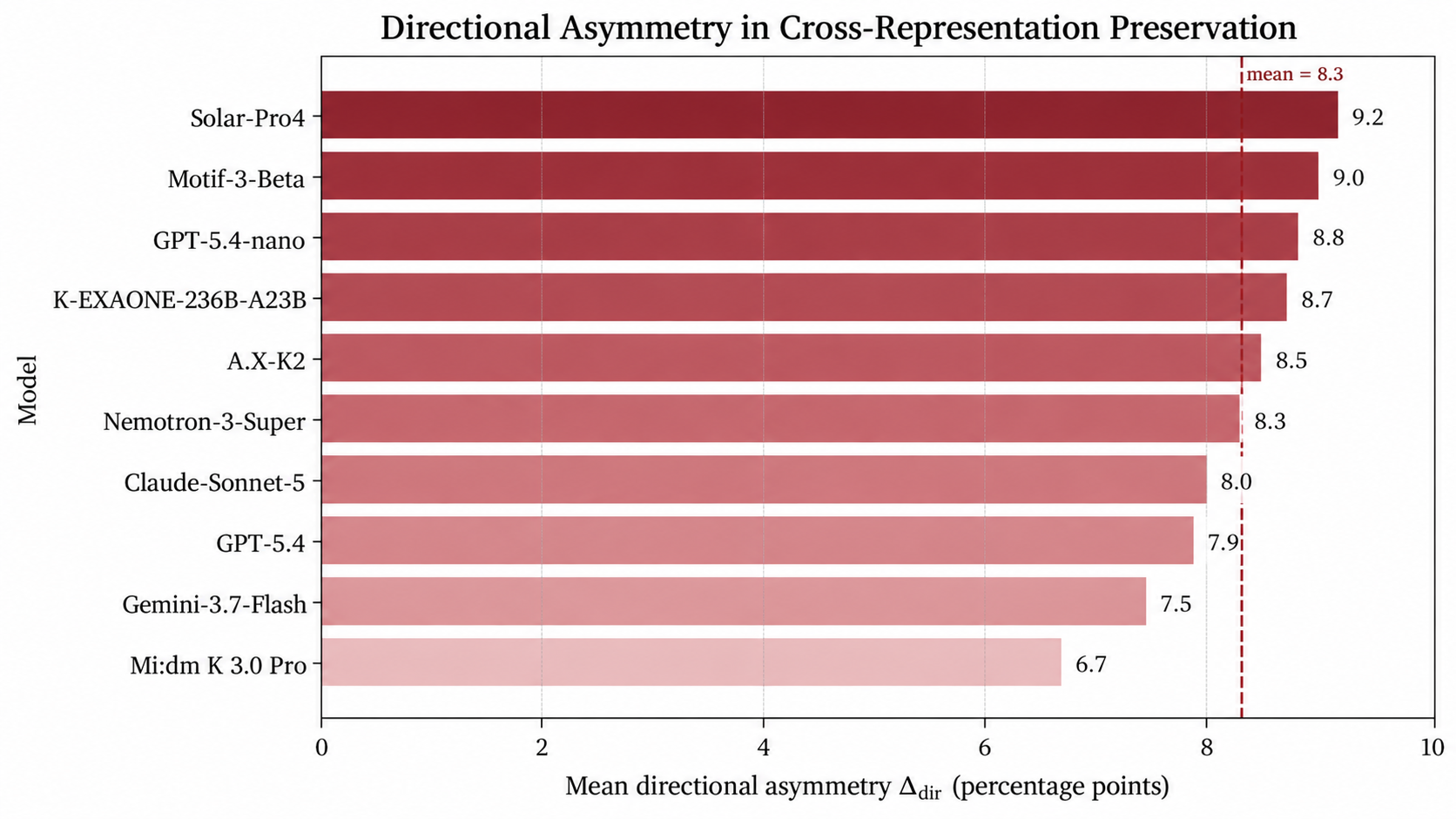}
    \caption{
    \textbf{Model-specific pooled source--target asymmetry.}
    $\Delta_{\mathrm{pool},FC}$ is the mean absolute Format Compliance
    difference between observed opposite transfers of the same representation
    pair. Opposite transfers may contain different artifact subsets.
    }
    \label{fig:directional-asymmetry}
\end{figure}

Figure~\ref{fig:directional-asymmetry} shows that pooled source--target
asymmetries recur across models. The JSON--Markdown pair provides a prominent
example: the observed transfer into Markdown has substantially higher Format
Compliance than the reverse transfer into JSON across multiple model
configurations. The magnitude of this pooled asymmetry also varies by model.

\subsection{Source- and Target-Side Profiles}
\label{app:source-target-difficulty}

The observed source--target variation can also be summarized according to
whether a representation appears on the input or output side. Let $x$ denote
an observed \textsc{Preserve} instance. We define source- and target-side
Format Compliance as
\[
S(s)
=
\mathbb{E}
\left[
FC(x)
\mid
\mathrm{source}(x)=s
\right],
\qquad
T(t)
=
\mathbb{E}
\left[
FC(x)
\mid
\mathrm{target}(x)=t
\right],
\]
where the expectations are taken over observed conversion instances. These
averages therefore reflect the unequal support reported in
Table~\ref{tab:preserve-support}.

The resulting source and target profiles are not equivalent. Markdown, for
example, has substantially higher observed performance as a target than as a
source, whereas JSON shows the opposite pattern. These profiles are consistent
with preservation performance varying with both recovery from the source
carrier and instantiation of the target contract; they should not be
interpreted as isolated source- or target-side causal effects.

\begin{figure}[t]
    \centering
    \includegraphics[width=\linewidth]
    {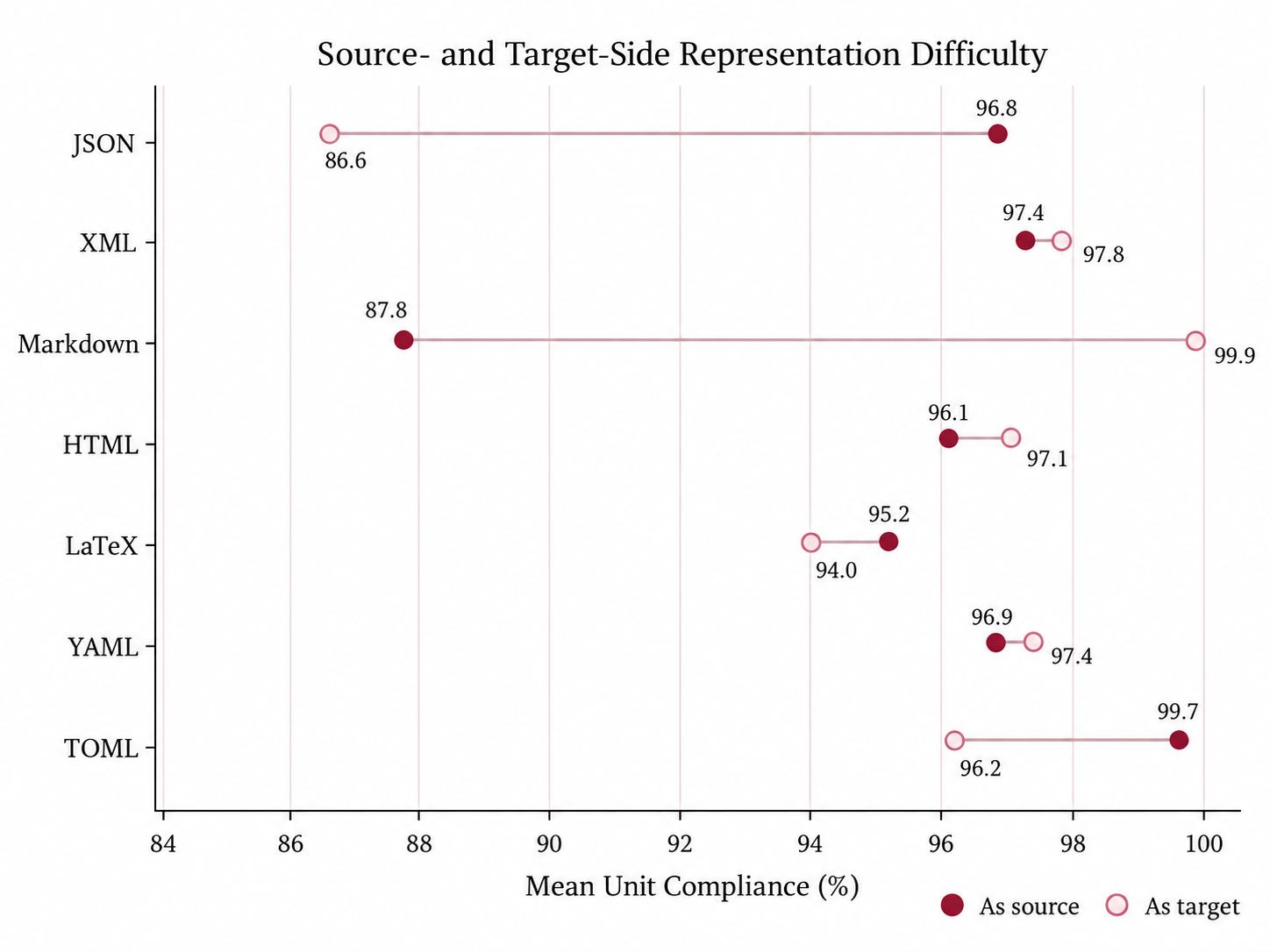}
    \caption{
    \textbf{Source- and target-side representation profiles.}
    Source scores average Format Compliance over observed outgoing transfers,
    and target scores average Format Compliance over observed incoming
    transfers.
    }
    \label{fig:source-target-difficulty}
\end{figure}

\subsection{Atomic Feature Performance}
\label{app:atomic-feature-performance}

\begin{figure*}[t]
    \centering
    \includegraphics[width=0.95\textwidth]
    {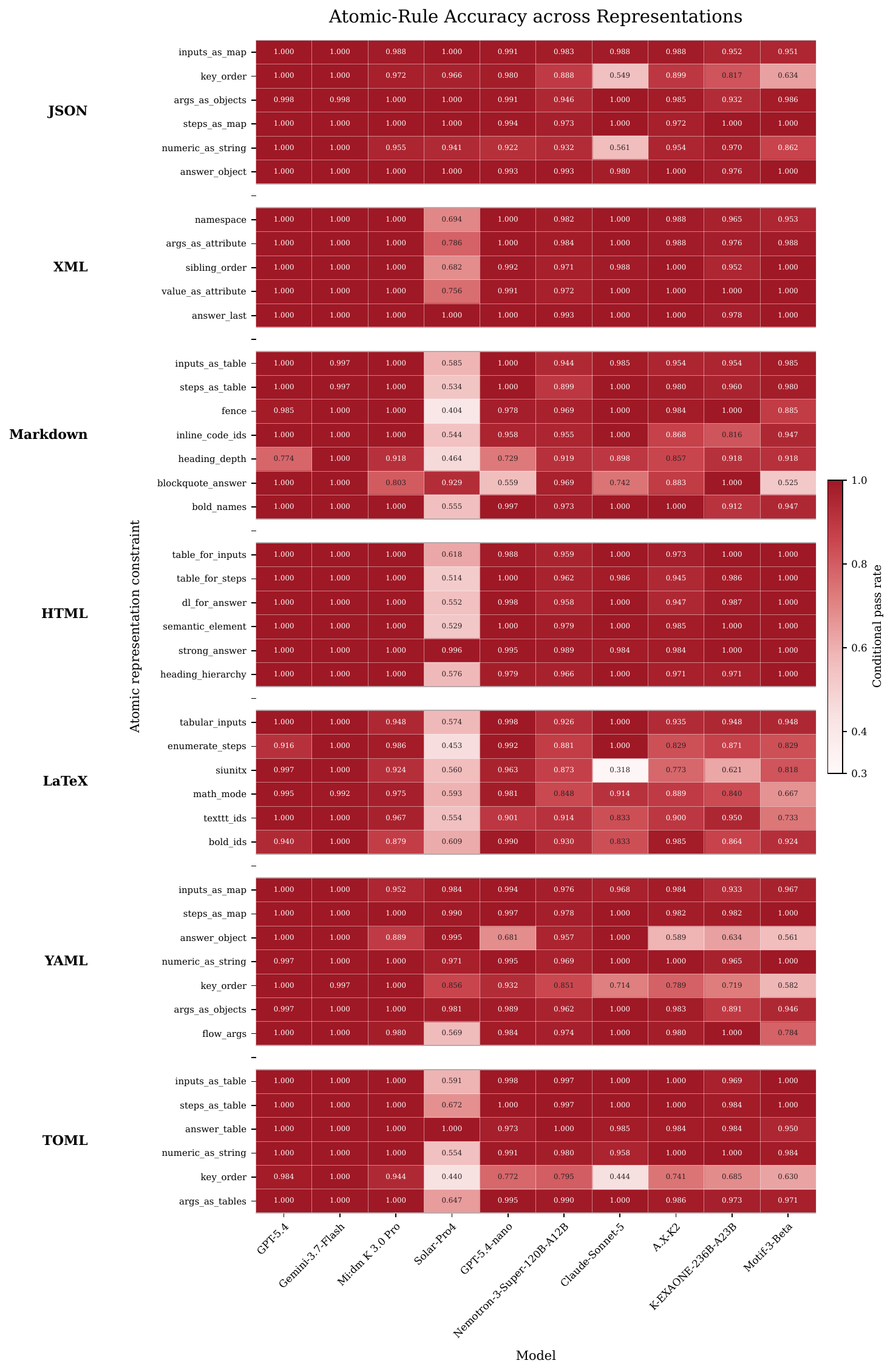}
    \caption{
    \textbf{Conditional pass rates for atomic representation constraints.}
    Each value is computed only over instances in which the corresponding
    rule is active. Rule labels additionally report the per-model activation
    count $n$; the same rule assignments are used for all model configurations.
    }
    \label{fig:atomic-feature-performance}
\end{figure*}

Figure~\ref{fig:atomic-feature-performance} reports the conditional pass rate
of each atomic rule together with its activation support. Support $n$ denotes
the number of unique \textsc{Realize} benchmark instances per model in which
the corresponding rule is active. Because rule assignments are fixed across
all ten model configurations, the support is identical across models and is
reported once per rule. Across L1, L2, and L3, the rule activations sum to
2,940 per model.

The realized support is not assumed to be uniform across atomic rules; the
reported counts make the sampling distribution explicit. Among the observed
activations, explicit ordering constraints recur among the lower-scoring
requirements across JSON, YAML, and TOML, while several challenging
\LaTeX{} rules involve native representation constructs such as mathematical
mode and identifier markup. Conditional pass rates should therefore be
interpreted together with their corresponding support.

\subsection{Atomic Failure Analysis}
\label{app:atomic-results}

The low conditional pass rates above do not all arise from the same failure
mechanism. We distinguish \emph{cascading failures}, where malformed output
prevents the requested structure from being recovered, from \emph{direct rule
violations}, where the document remains valid but the requested representation
constraint is not satisfied. Table~\ref{tab:atomic-failure-types} summarizes
representative difficult constraints together with their activation support.

\begin{table}[t]
\caption{
\textbf{Representative difficult atomic constraints.}
Mean accuracy is computed across the ten evaluated model configurations using
only \textsc{Realize} instances in which the corresponding rule is active.
Support $n$ denotes the number of active benchmark instances per model.
}
\centering
\small
\begin{tabular}{lrr}
\toprule
\textbf{Atomic rule} &
\textbf{Support $n$} &
\textbf{Mean Acc.} \\
\midrule

\texttt{toml.key\_order}
& 72 & 0.743 \\

\texttt{latex.siunitx}
& 69 & 0.785 \\

\texttt{yaml.answer\_object}
& 75 & 0.831 \\

\texttt{yaml.key\_order}
& 71 & 0.844 \\

\texttt{latex.math\_mode}
& 67 & 0.869 \\

\texttt{json.key\_order}
& 74 & 0.871 \\

\bottomrule
\end{tabular}
\label{tab:atomic-failure-types}
\end{table}

\paragraph{Cascading failure: TOML key ordering.}
The lowest observed mean pass rate,
\texttt{toml.key\_order}, illustrates how a representation-specific syntax
rule can amplify a local mistake. In TOML, a key written after a
\texttt{[[steps]]} header remains scoped to that table. Consequently, placing
the answer after the final step can remove the required root-level answer slot
even when the output appears visually reasonable.

\begin{PromptBlock}
[[steps]]
args = ["total_loss", "final_weight"]
id = "initial_weight"
op = "add"
value = "100"

answer = "100"
\end{PromptBlock}

Here, \texttt{answer} is parsed as a field of the final step rather than as a
top-level key. The resulting failure therefore propagates beyond
\texttt{key\_order} to other structural checks.

\paragraph{Direct violation: nested YAML key ordering.}
By contrast, \texttt{yaml.key\_order} frequently fails on otherwise valid
documents. A recurring pattern is that the model correctly orders the
top-level mapping but fails to propagate the ordering constraint to nested
mappings.

\begin{PromptBlock}
steps:
  - args:
      - {ref: input, name: jars}
      - {ref: input, name: cupcakes_per_jar}
    id: step1
    op: multiply
    value: 48
\end{PromptBlock}

The document is valid YAML, but the nested mappings place \texttt{ref} before
\texttt{name}, violating the requested alphabetical ordering. This example
shows that models can satisfy a global-looking constraint at the document
surface while failing to propagate it through nested structure.

Overall, the atomic analysis shows that low conditional pass rates arise from
both representation-specific syntactic fragility and failures to execute
fine-grained constraints. The same abstract requirement also shows substantial
carrier-specific variation within its observed support: mean
\texttt{key\_order} accuracy is 0.871 in JSON, 0.844 in YAML, and 0.743 in
TOML.

\subsection{Model-Level Performance Differences}
\label{app:model-profiles}

The preceding analyses show that no single ranking characterizes model
behavior across \textsc{Serial-7}. Table~\ref{tab:model-profiles} summarizes
prominent performance patterns using the main results together with the
representation- and preservation-level diagnostics.

\begin{table*}[t]
\caption{
\textbf{Model-level performance patterns across \textsc{Serial-7}.}
The table summarizes relative performance patterns rather than defining an
additional aggregate ranking.
}
\label{tab:model-profiles}
\centering
\small

\begin{tabular}{
@{}
p{0.19\textwidth}
p{0.36\textwidth}
p{0.36\textwidth}
@{}
}
\toprule
\textbf{Model} &
\textbf{Higher-performing setting} &
\textbf{Prominent pattern} \\
\midrule

GPT-5.4 &
High constrained-realization HSR with a small descriptive Realize--Preserve gap. &
Carrier-specific variation remains despite high aggregate performance. \\

Gemini-3.7-Flash &
Highest observed L3 HSR and high preservation, with similar performance
across the independently sampled L1--L3 conditions. &
A localized representation weakness remains visible, particularly on
\LaTeX{}. \\

Mi:dm &
High \textsc{Preserve} HSR and competitive REALIZE-L1 performance. &
HSR is noticeably lower in the independently sampled L3 condition than in
L1. \\

Solar-Pro4 &
High \textsc{Preserve} HSR and comparatively high HSR on selected carriers such as XML. &
Performance is less uniform across representation families. \\

GPT-5.4-nano &
High \textsc{Preserve} HSR despite substantially lower constrained-realization HSR than the top-performing models. &
HSR is substantially lower in the independently sampled L3 condition than in
L1. \\

Nemotron-3-Super &
Highest observed \textsc{Preserve} HSR among the evaluated models. &
Substantially lower L3 HSR, producing a large descriptive Realize--Preserve gap. \\

Claude-Sonnet-5 &
Highest observed unconstrained \textsc{Derive} accuracy. &
High \textsc{Derive} accuracy does not carry over uniformly to constrained
realization; L3 HSR is substantially lower. \\

A.X-K2 &
High \textsc{Derive} accuracy and substantially higher observed performance
under \textsc{Preserve}. &
Large descriptive separation between semantic accuracy and L3 realization. \\

K-EXAONE-236B-A23B &
High \textsc{Derive} accuracy. &
\textsc{Realize} and \textsc{Preserve} scores are substantially lower than its \textsc{Derive} accuracy. \\

Motif-3-Beta &
\textsc{Preserve} HSR is substantially higher when content is already
instantiated. &
Largest descriptive Realize--Preserve separation and lowest observed L3
realization HSR. \\

\bottomrule
\end{tabular}
\end{table*}

These profiles reinforce the central finding of the benchmark: semantic
accuracy, representation realization, and representation preservation expose
different failure modes. Reporting only a single structured-generation score would obscure whether
failure arises during content derivation, representation-contract realization,
multi-constraint generation, or artifact recovery from a source carrier.
\section{Evaluator Validation}
\label{app:evaluator-validation}

We validate the \textsc{Serial-7} evaluation harness through executable-verifier
audits, LLM-based cross-checks, and blinded human evaluation. Mechanical
verification is the primary evaluator for \textsc{Realize} and
\textsc{Preserve}, while \textsc{Derive} is evaluated with an LLM judge.

\subsection{Executable-Verifier Audits}
\label{app:mechanical-audits}

We audit the executable verification pipeline for renderer--parser consistency,
atomic-rule correctness, leakage, and prompt--checker alignment.

\begin{table}[H]
\caption{Executable-verifier audits for \textsc{Serial-7}.}
\centering
\small
\begin{tabular}{
@{}
>{\raggedright\arraybackslash}p{0.23\linewidth}
>{\raggedright\arraybackslash}p{0.48\linewidth}
>{\raggedright\arraybackslash}p{0.20\linewidth}
@{}
}
\toprule
\textbf{Audit} &
\textbf{Validation target} &
\textbf{Result} \\
\midrule

Round-trip validation &
Whether gold realizations can be serialized, recovered as Canonical IR, and
accepted by the corresponding predicates. &
Base: 1,239/1,239;
\textsc{Realize}: 1,470/1,470;
\textsc{Preserve}: 1,695/1,695. \\

Atomic-rule validation &
Whether sampled rule combinations are recoverable and near-miss examples are
rejected by the intended checker. &
777 combinations, 0 failures;
43/43 rules covered by near-miss tests. \\

Leakage audit &
Whether valid transformations are incorrectly penalized and format predicates
remain invariant to semantics-preserving perturbations. &
987 valid transformations, 0 false penalties;
29/33 rule types invariant. \\

Prompt--checker alignment &
Whether scored predicates are stated and checked consistently across Korean
and English specifications. &
0 unstated checks;
0 unmeasured requirements;
0 language-symmetry violations. \\

\bottomrule
\end{tabular}
\label{tab:mechanical-audits}
\end{table}

The four semantics-sensitive rule types are
\texttt{answer\_last}, \texttt{answer\_object},
\texttt{blockquote\_answer}, and \texttt{strong\_answer}, whose definitions
explicitly refer to the answer representation or location.

\subsection{LLM-Judge Cross-Checks}
\label{app:llm-judge-agreement}

All LLM judgments use
\texttt{google/gemini-3.1-pro-preview} through OpenRouter with temperature
0.0 and a fixed rubric. For \textsc{Realize} and \textsc{Preserve}, the judge
cross-checks sampled atomic constraints without access to the executable
verdict or gold serialization; mechanical verification remains primary.

For \textsc{Derive}, the judge receives the question, reference answer, and
model response and evaluates the final committed answer. A deterministic
post-check requires the judged answer to be recoverable from the response.

\subsection{Human Validation}
\label{app:human-validation}

Two annotators independently label 300 outputs while blinded to automatic
verdicts. The structured sample contains 210 outputs covering seven formats,
three constraint-count conditions, and two languages
($7\times3\times2\times5$); the 90 \textsc{Derive} outputs are stratified
across language, model, and LLM-judge verdict. Structured outputs are judged
only against explicitly stated representation requirements, and disagreements
are adjudicated to obtain final labels.

\begin{table}[H]
\caption{
\textbf{Human validation of \textsc{Serial-7}.}
Human--Human reports agreement and Cohen's $\kappa$ before adjudication;
Automatic--Human compares the automatic evaluator with final human labels.
}
\centering
\small
\setlength{\tabcolsep}{4pt}
\begin{tabularx}{\linewidth}{
@{}
>{\raggedright\arraybackslash}X
>{\centering\arraybackslash}p{0.055\linewidth}
>{\raggedright\arraybackslash}X
>{\centering\arraybackslash}p{0.19\linewidth}
>{\centering\arraybackslash}p{0.16\linewidth}
@{}
}
\toprule
\textbf{Target} &
\textbf{$n$} &
\textbf{Sampling} &
\textbf{Human--Human} &
\textbf{Automatic--Human} \\
\midrule

Active-constraint compliance &
210 &
$7$ formats $\times$ $3$ conditions $\times$ $2$ languages $\times$
$5$ examples &
203/210 (96.7\%);
$\kappa=0.93$ &
201/210 (95.7\%) \\

Overall format compliance &
210 &
Same structured sample &
204/210 (97.1\%);
$\kappa=0.94$ &
202/210 (96.2\%) \\

\textsc{Derive} semantic correctness &
90 &
Stratified by language, model, and judge verdict &
89/90 (98.9\%);
$\kappa=0.97$ &
88/90 (97.8\%) \\

\bottomrule
\end{tabularx}
\label{tab:human-evaluation}
\end{table}

The results provide high aggregate evaluator agreement, but the structured
sample is designed primarily for broad coverage rather than class-conditional
error estimation. We therefore perform an additional verdict-balanced audit.

\paragraph{Targeted verifier audit.}
We sample 120 structured outputs across the ten model configurations,
including 60 automatic-pass and 60 automatic-fail cases. Annotators remain
blind to the verifier verdict.

\begin{table}[H]
\centering
\small
\caption{
\textbf{Targeted executable-verifier audit.}
Rows denote verifier decisions and columns denote final human labels.
}
\begin{tabular}{lcc}
\toprule
& \multicolumn{2}{c}{\textbf{Human judgment}} \\
\cmidrule(lr){2-3}
\textbf{Executable verifier} &
\textbf{Pass} &
\textbf{Fail} \\
\midrule
Pass & 58 & 2 \\
Fail & 4 & 56 \\
\bottomrule
\end{tabular}
\label{tab:verifier-confusion}
\end{table}

The verifier agrees with human judgments on 114/120 outputs (95.0\%).
Human review rejects 2/60 automatic-pass cases (3.3\%) and accepts 4/60
automatic-fail cases (6.7\%). These results characterize verifier error
direction while avoiding the interpretation of aggregate agreement as
evidence that small between-model score differences represent resolved
rankings.
\section{Experimental Configuration}
\label{app:experimental-configuration}

This appendix describes the inference configurations used for the SERIAL-7
evaluation. Evaluator-side LLM judges are described separately in
Appendix~\ref{app:evaluator-validation}. The primary evaluation reported in
this paper comprises 3,489 prompts across \textsc{Derive}, \textsc{Realize},
and \textsc{Preserve}: 324, 1,470, and 1,695 prompts, respectively.

We evaluate ten model configurations. GPT-5.4, GPT-5.4-nano,
Gemini-3.7-Flash, Claude-Sonnet-5, Nemotron-3-Super-120B-A12B, and
Solar-Pro4 are accessed through OpenRouter. Motif-3-Beta, A.X-K2,
K-EXAONE-236B-A23B, and Mi:dm K 3.0 Pro are hosted locally using vLLM.
All prompts are represented as single user messages without an additional
system prompt and are serialized using the corresponding model-specific
chat template.

\subsection{API-Served Models}
\label{app:api-models}

All API-served models are accessed through OpenRouter. Requests use a
temperature of 0.0 and explicitly request reasoning through
\texttt{reasoning.enabled=true}. The maximum generation length is 12,000
tokens for all API-served models except Claude-Sonnet-5, for which it is
16,000 tokens. Table~\ref{tab:api-models} reports the exact model identifiers
and observed provider routes.

\begin{table}[H]
\caption{Configurations of the API-served models used for SERIAL-7.}
\centering
\scriptsize
\setlength{\tabcolsep}{3.5pt}
\begin{tabular}{
@{}
>{\raggedright\arraybackslash}p{0.20\linewidth}
>{\raggedright\arraybackslash}p{0.34\linewidth}
>{\raggedright\arraybackslash}p{0.27\linewidth}
>{\centering\arraybackslash}p{0.11\linewidth}
@{}
}
\toprule
\textbf{Model} &
\textbf{OpenRouter identifier} &
\textbf{Observed provider} &
\textbf{Max. tokens} \\
\midrule

GPT-5.4 &
\texttt{openai/gpt-5.4} &
OpenAI &
12,000 \\

GPT-5.4-nano &
\texttt{openai/gpt-5.4-nano} &
OpenAI &
12,000 \\

Gemini-3.7-Flash &
\texttt{google/gemini-3.7-flash} &
Google; Google AI Studio &
12,000 \\

Claude-Sonnet-5 &
\texttt{anthropic/claude-sonnet-5} &
Claude Platform on AWS &
16,000 \\

Nemotron-3-Super-120B-A12B &
\texttt{nvidia/nemotron-3-super-120b-a12b} &
DeepInfra; DigitalOcean &
12,000 \\

Solar-Pro4 &
\texttt{upstage/solar-pro4} &
Upstage &
12,000 \\

\bottomrule
\end{tabular}
\label{tab:api-models}
\end{table}

Reasoning is requested for every API call, although an explicit reasoning
trace is not returned for every request because its availability depends on
the model and provider. When a reasoning trace is returned separately from
the final response, it is removed before evaluation. SERIAL-7 scores are
computed only from the final response.

\subsection{Locally Hosted Models and Serving Configuration}
\label{app:models-serving}

The four locally hosted models are Motif-3-Beta, A.X-K2,
K-EXAONE-236B-A23B, and Mi:dm K 3.0 Pro.
Table~\ref{tab:models-serving} summarizes their weight representations and
software environments. Runtime parallelization and decoding settings are
reported separately in Table~\ref{tab:model-inference}.

\begin{table}[H]
\caption{Weight and software configurations of the locally hosted models.}
\centering
\small
\begin{tabular}{
@{}
>{\raggedright\arraybackslash}p{0.23\linewidth}
>{\raggedright\arraybackslash}p{0.28\linewidth}
>{\centering\arraybackslash}p{0.21\linewidth}
>{\centering\arraybackslash}p{0.20\linewidth}
@{}
}
\toprule
\textbf{Model} &
\textbf{Weight representation} &
\textbf{vLLM} &
\textbf{Transformers} \\
\midrule

Motif-3-Beta &
ModelOpt block FP8 &
\texttt{0.20.2+motif3.g0b6f55172} &
\texttt{5.14.1} \\

A.X-K2 &
FP8 E4M3, block $128 \times 128$ &
\texttt{0.23.0} &
\texttt{5.18.0.dev0} \\

K-EXAONE-236B-A23B &
BF16 &
\texttt{0.20.2} &
\texttt{5.8.1} \\

Mi:dm K 3.0 Pro &
BF16; no explicit quantization override &
\texttt{0.17.0} &
\texttt{4.57.6} \\

\bottomrule
\end{tabular}
\label{tab:models-serving}
\end{table}

Motif-3-Beta uses \texttt{modelopt\_blockfp8} weights and is executed with the
dedicated container
\texttt{ghcr.io/motiftechnologies/vllm:v0.20.2-motif3.rc2}.
A.X-K2 uses FP8 quantization automatically detected from the model
configuration. K-EXAONE-236B-A23B uses BF16 weights loaded from safetensors,
while Mi:dm K 3.0 Pro is served without an explicit quantization override.

\subsection{Local Inference and Reasoning Settings}
\label{app:inference-settings}

All four locally hosted models are evaluated using offline batched generation
through vLLM's \texttt{LLM.generate()} interface. We use a maximum generation
length of 32,768 tokens, BF16 runtime computation, and a fixed random seed of
42. Motif-3-Beta and A.X-K2 use FP8-quantized weights, while K-EXAONE-236B-A23B
and Mi:dm K 3.0 Pro use BF16 weights. For models with an explicit reasoning
control, we set \texttt{enable\_thinking=True}. Reasoning traces enclosed in
\texttt{<think>...</think>} are removed before evaluation.

Sampling follows each model's recommended configuration. Motif-3-Beta and
K-EXAONE-236B-A23B use temperature 1.0, while A.X-K2 and Mi:dm K 3.0 Pro use
temperature 0.6. All four local models use \texttt{top\_p=0.95}. Other serving
settings are summarized in Table~\ref{tab:model-inference}. API-hosted models
are evaluated separately with temperature 0.0.

\begin{table}[H]
\caption{Inference, sampling, and parallelization settings for the locally
hosted models.}
\centering
\small
\resizebox{\textwidth}{!}{
\begin{tabular}{@{}lcccc@{}}
\toprule
\textbf{Parameter} &
\textbf{Motif-3-Beta} &
\textbf{A.X-K2} &
\textbf{K-EXAONE-236B-A23B} &
\textbf{Mi:dm K 3.0 Pro} \\
\midrule

Maximum context length &
262,144 &
131,072 &
131,072 &
131,072 \\

Temperature &
1.0 &
0.6 &
1.0 &
0.6 \\

Top-$p$ &
0.95 &
0.95 &
0.95 &
0.95 \\

GPU-memory utilization &
0.85 &
0.90 &
0.90 &
0.90 \\

Tensor parallel size &
1 &
8 &
4 &
8 \\

Data parallel size &
8 &
1 &
1 &
1 \\

Pipeline parallel size &
1 &
1 &
1 &
1 \\

Speculative decoding &
Self, 1 token &
Off &
Off &
Off \\

Expert parallelism &
On &
Off &
On &
Off \\

EP weight filtering &
Off &
Off &
On &
Off \\

\bottomrule
\end{tabular}
}
\label{tab:model-inference}
\end{table}

Motif-3-Beta runs on eight H200 GPUs with DP8~$\times$~TP1, one-token
self-speculative decoding, and expert parallelism. A.X-K2 runs with TP8 across
eight GPUs. K-EXAONE-236B-A23B runs with TP4 across four GPUs, with expert
parallelism and expert-parallel weight filtering enabled. Mi:dm K 3.0 Pro runs
with TP8 across eight GPUs.

\subsection{Compute Environment}
\label{app:compute-environment}

All locally hosted experiments are executed on a single server equipped with
eight NVIDIA H200 GPUs. Each GPU provides 143,771\,MiB of device memory, for
approximately 1.1\,TiB of aggregate GPU memory. The eight GPUs are fully
interconnected through NVLink (\texttt{NV18}). The host system contains two
Intel Xeon Platinum 8480C processors with 48 cores per socket and 1.8\,TiB of
system memory.

The common software environment uses NVIDIA driver 580.126.20, CUDA 13.0,
and PyTorch 2.11.0+cu130. Motif-3-Beta, A.X-K2, and Mi:dm K 3.0 Pro use all
eight available H200 GPUs, whereas K-EXAONE-236B-A23B uses four GPUs.
Model-specific vLLM and Transformers versions are reported in
Table~\ref{tab:models-serving}.
\section{Benchmark Construction Details}
\label{app:benchmark_construction}

SERIAL-7 follows a controlled-comparison design in which heterogeneous source
problems are normalized into a representation-independent Canonical IR and
then reused across language and representation conditions. This construction
keeps the underlying semantic artifact fixed while varying only its surface
realization.

\subsection{Source Sampling and Normalization}
\label{app:source_sampling}

We sample 500 seed instances from four public reasoning benchmarks using
benchmark-specific strata designed to preserve diversity in problem structure
and reasoning requirements.

\begin{table}[H]
\caption{Source benchmarks and sampling strata used to construct the initial
seed set.}
\centering
\small
\begin{tabular}{
@{}
>{\raggedright\arraybackslash}p{0.20\linewidth}
>{\raggedright\arraybackslash}p{0.34\linewidth}
>{\raggedright\arraybackslash}p{0.36\linewidth}
@{}
}
\toprule
\textbf{Source} &
\textbf{Problem structure} &
\textbf{Sampling stratum} \\
\midrule

TabMWP &
Table-grounded arithmetic &
Question type, table scope, estimated reasoning depth, and operation type. \\

HRM8K &
Korean mathematical reasoning &
Predefined benchmark subset. \\

SciBench &
Scientific calculation &
Scientific subject. \\

MMLU-Pro (STEM) &
STEM reasoning &
Subject. \\

\bottomrule
\end{tabular}
\label{tab:source-benchmarks}
\end{table}

Each selected item is normalized into a shared artifact containing the
question and supporting materials, semantic inputs, an optional verified
computation graph, and a normalized final answer. Source provenance is retained
during construction to verify that normalized values remain grounded in the
original benchmark record.

Normalization includes three principal checks. First, the normalized answer
must be consistent with the official benchmark answer under the source
dataset's answer convention. Second, all required inputs and supporting values
must be traceable to the source instance without introducing unsupported
information. Third, when a computation graph is available, its operations,
dependencies, and intermediate values must be executable and consistent with
the final answer.

Items with a valid problem and final answer but without a machine-verifiable
computation graph are retained for answer-level evaluation only. Evaluation
components that require intermediate computation structure are restricted to
artifacts with a verified graph.

\subsection{Gold-Set Validation}
\label{app:gold_validation}

Gold-set selection depends only on the source problem and its normalized
artifact; outputs from evaluated models are never used for retention or
filtering. Candidate artifacts are retained only when they satisfy the
following criteria:

\begin{table}[H]
\caption{Principal criteria used for gold-set validation.}
\centering
\small
\begin{tabular}{
@{}
>{\raggedright\arraybackslash}p{0.27\linewidth}
>{\raggedright\arraybackslash}p{0.63\linewidth}
@{}
}
\toprule
\textbf{Criterion} &
\textbf{Requirement} \\
\midrule

Source validity &
The original problem, supporting materials, and normalized artifact must be
well formed and mutually consistent. \\

Answer consistency &
The normalized final answer must be reconcilable with the official benchmark
answer. \\

Grounding and leakage &
Required values must be supported by the source record, and auxiliary text
must not expose the answer without legitimate grounding. \\

Representation compatibility &
Normalized values must be representable under the supported serialization
contracts without introducing construction artifacts into the question. \\

Redundancy control &
Repeated or near-duplicate question templates are canonicalized and capped to
avoid over-representation of a single template family. \\

\bottomrule
\end{tabular}
\label{tab:gold-validation-criteria}
\end{table}

To make the filtering process explicit, we assign each rejected candidate to
the first validation gate that remains unsatisfied after any repairable issue
has been corrected. The resulting exclusion categories are therefore mutually
exclusive.
\begin{table}[t]
\centering
\small
\setlength{\tabcolsep}{5pt}
\renewcommand{\arraystretch}{1.08}
\caption{Gold-set construction and retention statistics for
\textsc{SERIAL-7}. Rejection categories are mutually exclusive and
assigned to the first unresolved validation failure after repair.}
\label{tab:gold_filtering}

\begin{tabular}{lrr}
\toprule
\multicolumn{3}{c}{\textbf{(a) Filtering flow}} \\
\midrule
\textbf{Stage} & \textbf{Excluded} & \textbf{Remaining} \\
\midrule
Initial seed set                     & -- & 500 \\
Source / normalization validity      & 72 & 428 \\
Answer inconsistency                 & 61 & 367 \\
Grounding or leakage failure         & 54 & 313 \\
Representation incompatibility       & 83 & 230 \\
Redundancy / duplicate-cap filtering & 53 & 177 \\
\midrule
\textbf{Total excluded}              & \textbf{323} & \textbf{177} \\
\bottomrule
\end{tabular}

\vspace{0.8em}

\begin{tabular}{lrrr}
\toprule
\multicolumn{4}{c}{\textbf{(b) Source-wise retention}} \\
\midrule
\textbf{Source} & \textbf{Initial} & \textbf{Retained}
& \textbf{Retention (\%)} \\
\midrule
TabMWP          & 200 & 85 & 42.5 \\
HRM8K           & 120 & 47 & 39.2 \\
SciBench        & 100 & 25 & 25.0 \\
MMLU-Pro (STEM) & 80  & 20 & 25.0 \\
\midrule
\textbf{Total}  & \textbf{500} & \textbf{177} & \textbf{35.4} \\
\bottomrule
\end{tabular}
\end{table}

Thus, 177 of the initial 500 seed instances remain as validated semantic
artifacts. Among these retained artifacts, 37 contain a mechanically verified
computation graph and 140 are retained as answer-only artifacts. Lack of a
verified graph is therefore not itself treated as an exclusion criterion.
Canonical renderings are additionally checked through
serialization--extraction round trips; the corresponding evaluator audits are
reported in Appendix~\ref{app:evaluator-validation}.

\paragraph{Filtering and retention.}
We initially sampled 500 source problems and retained 177 gold artifacts after
normalization, verification, and redundancy control. The remaining 323 items
were excluded by deterministic construction-time gates rather than by model
performance. Specifically, items could be removed because of source parsing
failure, unresolved answer verification, insufficient semantic grounding,
incompatibility with the representation contracts, or redundancy control.
Table~\ref{tab:gold_filtering} reports the number of items removed
at each stage and the resulting retained set.

Because representation compatibility is itself one of the filtering criteria,
the final benchmark should not be interpreted as an unbiased sample of the
original source distributions. In particular, the retained set may favor
artifacts whose semantics can be represented reliably under the benchmark
contracts. We therefore report the complete filtering flow and source-wise
retention statistics below. No evaluated model output was used at any stage
of benchmark filtering or retention.

\subsection{Bilingual Pair Construction}
\label{app:bilingual_construction}

Korean and English instances are generated from the same validated semantic
artifact rather than normalized independently. This design keeps answers,
required quantities, and computation structure fixed while allowing the
question and supporting text to be expressed naturally in each language.

\begin{table}[H]
\caption{Controlled construction of Korean--English benchmark pairs.}
\centering
\small
\begin{tabular}{
@{}
>{\raggedright\arraybackslash}p{0.28\linewidth}
>{\raggedright\arraybackslash}p{0.25\linewidth}
>{\raggedright\arraybackslash}p{0.38\linewidth}
@{}
}
\toprule
\textbf{Component} &
\textbf{Across KO--EN pair} &
\textbf{Validation} \\
\midrule

Question and source text &
Language-specific &
Check internal language consistency. \\

Answer and numeric values &
Semantically fixed &
Require equivalent answers and preservation of required quantities. \\

Computation graph &
Fixed &
Reuse the same operations, dependencies, and constants. \\

Grounding evidence &
Language-specific &
Verify that required inputs remain supported in the corresponding
language-specific source text. \\

\bottomrule
\end{tabular}
\label{tab:bilingual-pair-construction}
\end{table}

\subsubsection{Example Korean--English Pair}
\label{app:bilingual-example}

Table~\ref{tab:bilingual-example} shows a paired realization of the same
validated artifact. The question and supporting materials are
language-specific, while the answer and computation remain shared across
the two language arms.

\begin{table}[H]
\caption{Example bilingual pair constructed from the same semantic artifact.
Question and material surfaces vary by language, while the answer, semantic
input values, and computation graph are shared.}
\centering
\small
\renewcommand{\arraystretch}{1.25}
\begin{tabular}{
@{}
>{\raggedright\arraybackslash}p{0.15\linewidth}
>{\raggedright\arraybackslash}p{0.38\linewidth}
>{\raggedright\arraybackslash}p{0.38\linewidth}
@{}
}
\toprule
&
\textbf{Korean arm} &
\textbf{English arm} \\
\midrule

Question &
Emily가 7개의 Caribbean 휴가 패키지를 사려면 돈이 얼마나 필요합니까? &
How much money does Emily need to buy 7 Caribbean vacation packages? \\

\addlinespace

Materials &
Caribbean 휴가 패키지 $\mid$ \$503 \newline
Australian 휴가 패키지 $\mid$ \$554 \newline
Hawaiian 휴가 패키지 $\mid$ \$613 \newline
European 휴가 패키지 $\mid$ \$867 &
Caribbean vacation package $\mid$ \$503 \newline
Australian vacation package $\mid$ \$554 \newline
Hawaiian vacation package $\mid$ \$613 \newline
European vacation package $\mid$ \$867 \\

\addlinespace

Shared answer &
\multicolumn{2}{c}{$3521$} \\

Shared computation &
\multicolumn{2}{c}{
\texttt{R1} $=$
\texttt{multiply}(\texttt{caribbean\_price}, $7$) $=3521$
} \\

\bottomrule
\end{tabular}
\label{tab:bilingual-example}
\end{table}

Of the 177 validated artifacts, 162 have validated realizations in both
languages. All direct Korean--English comparisons are restricted to these
paired items.

\subsection{Canonical IR and Representation Rendering}
\label{app:canonical_example}

After construction-time provenance fields are removed, each validated artifact
is projected to the scored Canonical IR
\[
z=(v,I,S),
\]
where $v$ is the final answer,
\[
I=[(k_1,v_1),\ldots,(k_n,v_n)]
\]
is the ordered sequence of input identifier--value pairs, and
\[
S=[s_1,\ldots,s_m]
\]
is the optional ordered sequence of computation-step records. Each step record
contains its identifier, operator, ordered argument references, and value.
These identifiers and sequence orders are part of the scored artifact rather
than incidental serialization details.

Given a representation contract $c$, the corresponding realization is
obtained by
\[
y_c = R_c(z).
\]
The renderer changes only the carrier-specific surface form while preserving
the Canonical IR and its required artifact invariants.

For example, a TabMWP instance requiring the total cost of seven \$503
vacation packages is represented by the following shared artifact.

\begin{table}[H]
\caption{Example Canonical IR shared across representation conditions.}
\centering
\small
\begin{tabular}{
@{}
>{\raggedright\arraybackslash}p{0.20\linewidth}
>{\raggedright\arraybackslash}p{0.68\linewidth}
@{}
}
\toprule
\textbf{IR component} &
\textbf{Canonical content} \\
\midrule

Answer &
$3521$ \\

Input &
\texttt{caribbean\_price} $= 503$ \\

Step &
\texttt{R1} $=$
\texttt{multiply}(\texttt{caribbean\_price}, $7$) $= 3521$ \\

\bottomrule
\end{tabular}
\label{tab:canonical-example}
\end{table}

The seven output families differ only in the deterministic rendering
$R_c$ applied to the shared artifact. Their base contracts and atomic
representation constraints are specified in Appendix~\ref{app:prompts}.
\section{Prompts and Representation Contracts}
\label{app:prompts}

This appendix documents the evaluation prompts and representation contracts
used in SERIAL-7. Benchmark construction first produces a validated,
representation-independent Canonical IR, as described in
Appendix~\ref{app:benchmark_construction}. The evaluation conditions differ
in the model input, required operation, and representation requirements:
\textsc{Derive} answers without a prescribed representation,
\textsc{Realize} derives and renders an answer under an explicit representation
contract with sampled atomic constraints, and \textsc{Preserve} converts an
already instantiated document under the base target representation contract
without recomputing its content or activating sampled atomic constraints.

All evaluation prompts are serialized as a single user message using the
model-specific chat template; no additional system prompt is used.

\subsection{Evaluation Prompt Organization}
\label{app:prompt-organization}

\begin{table}[H]
\caption{Evaluation prompt organization in SERIAL-7.}
\centering
\small
\begin{tabular}{
@{}
>{\raggedright\arraybackslash}p{0.16\linewidth}
>{\raggedright\arraybackslash}p{0.27\linewidth}
>{\raggedright\arraybackslash}p{0.28\linewidth}
>{\raggedright\arraybackslash}p{0.20\linewidth}
@{}
}
\toprule
\textbf{Condition} &
\textbf{Input} &
\textbf{Required behavior} &
\textbf{Representation requirement} \\
\midrule

\textsc{Derive} &
Question and optional materials &
Solve the problem and provide the final answer. &
None \\

\textsc{Realize} &
Question and optional materials &
Solve the problem and express the result under a target contract. &
Base target representation + 1--3 sampled atomic constraints \\

\textsc{Preserve} &
Instantiated source document &
Rewrite the same semantic content in a target representation. &
Base target representation only; no sampled atomic constraints \\

\bottomrule
\end{tabular}
\label{tab:prompt-organization}
\end{table}

For \textsc{Realize}, the representation-specific portion of the prompt is
assembled as
\[
\begin{aligned}
\textsc{Instructions}_{f}
&= \textsc{AnswerRealization}_{f} \\
&\quad \oplus
[\textsc{GraphSyntax}_{f}\oplus\textsc{GraphValidity}] \\
&\quad \oplus
[\textsc{AtomicClauses}],
\end{aligned}
\]
where graph-related terms are present only for graph-bearing artifacts.
For the evaluated L1, L2, and L3 conditions, one, two, and three sampled
atomic clauses are included, respectively.

\subsection{Derive: Unconstrained Answering}
\label{app:derive-prompt}

\textsc{Derive} imposes no carrier, serialization syntax, schema, field
organization, ordering rule, or atomic representation constraint.

\PromptTitle{Derive Prompt}
\begin{PromptBlock}
Answer the question. You may reply in any form you like.
\end{PromptBlock}

Because \textsc{Derive} does not specify a deterministic answer slot, it is
evaluated only for final-answer correctness using the semantic evaluator
described in Appendix~\ref{app:evaluator-validation}.

\subsection{Realize: Constrained Representation}
\label{app:realize-prompts}

\textsc{Realize} requires the model to solve the problem and serialize the
result under an explicit target representation. Each request is assembled as
a single user message with the following structure.

\PromptTitle{Realize: Assembled User Message}
\begin{PromptBlock}
## Materials
{materials}

## Question
{question}

## Instructions
{answer realization for target format}
{format-specific graph syntax and graph-validity requirements, if applicable}
{sampled atomic representation constraints}

## Constraints
{common content and output constraints}
\end{PromptBlock}

The \texttt{Materials} section is omitted when the source instance contains no
separate supporting material.

\begin{table}[H]
\caption{Components of the \textsc{Realize} prompt.}
\centering
\small
\begin{tabular}{
@{}
>{\raggedright\arraybackslash}p{0.26\linewidth}
>{\raggedright\arraybackslash}p{0.25\linewidth}
>{\raggedright\arraybackslash}p{0.40\linewidth}
@{}
}
\toprule
\textbf{Component} &
\textbf{When used} &
\textbf{Function} \\
\midrule

Answer realization &
All \textsc{Realize} instances &
Defines the format-specific answer slot. \\

Format-specific graph syntax &
Graph-bearing instances &
Defines how inputs, steps, references, and values are serialized. \\

Shared graph validity &
Graph-bearing instances &
Constrains operation validity, dependency order, and consistency with the
final answer. \\

Atomic representation constraints &
All \textsc{Realize} L1--L3 instances &
Modify local aspects of the base representation while preserving the same
semantic artifact. \\

Common constraints &
All constrained instances &
Prevent unsupported additions and restrict the output to the requested
document. \\

\bottomrule
\end{tabular}
\label{tab:realize-components}
\end{table}

\subsubsection{Base Representation Contracts}
\label{app:base-representation-contracts}

The seven representation families share the same Canonical IR but expose
different carrier-specific answer slots and graph structures.
Table~\ref{tab:base-contracts} summarizes the base contracts used in
\textsc{Realize} and reused as target contracts in \textsc{Preserve}.

\begin{table*}[t]
\centering
\small
\setlength{\tabcolsep}{4pt}
\renewcommand{\arraystretch}{1.15}

\caption{Base representation contracts used in \textsc{Realize} and as target
contracts in \textsc{Preserve}.}
\label{tab:base-contracts}

\begin{tabular}{
@{}
>{\raggedright\arraybackslash}p{0.10\textwidth}
>{\raggedright\arraybackslash}p{0.35\textwidth}
>{\raggedright\arraybackslash}p{0.47\textwidth}
@{}
}
\toprule
\textbf{Format} &
\textbf{Answer realization} &
\textbf{Inputs and steps} \\
\midrule

JSON &
A root JSON object containing \texttt{answer}. &
\texttt{inputs} and \texttt{steps} are arrays of structured objects containing
the corresponding names, identifiers, operations, arguments, and values. \\

\addlinespace

XML &
Exactly one \texttt{<response>} root containing one
\texttt{<answer>} element. &
Inputs appear under \texttt{<inputs>} and steps under \texttt{<steps>};
references are represented by format-specific elements and attributes. \\

\addlinespace

Markdown &
The document begins with \texttt{\# Result} and an
\texttt{- Answer:} field. &
\texttt{\#\# Inputs} and \texttt{\#\# Steps} contain ordered records with
nested fields for values, operations, and arguments. \\

\addlinespace

HTML &
Exactly one \texttt{<main>} containing
\texttt{<output id="answer">}. &
Inputs and steps are represented under elements identified by
\texttt{inputs} and \texttt{steps}, with record values encoded using
carrier-specific attributes. \\

\addlinespace

LaTeX &
Exactly one \texttt{response} environment containing
\texttt{\textbackslash Answer\{...\}}. &
Inputs and steps are represented by dedicated environments and
\texttt{\textbackslash Input} and \texttt{\textbackslash Step} macros. \\

\addlinespace

YAML &
A root mapping containing \texttt{answer}. &
\texttt{inputs} and \texttt{steps} are represented as structured YAML
collections. \\

\addlinespace

TOML &
A top-level \texttt{answer} key. &
Inputs and steps are represented using TOML array tables
\texttt{[[inputs]]} and \texttt{[[steps]]}. \\

\bottomrule
\end{tabular}
\end{table*}

\subsubsection{Shared Graph Validity}
\label{app:shared-graph-contract}

When a verified computation graph is available, all representation families
share the same representation-independent graph requirements:

\begin{itemize}
    \item input names and step identifiers must be unique;
    \item operations must belong to the supported arithmetic operation set;
    \item quantities provided by the problem are represented as inputs, while
          earlier computed results are referenced by step identifier;
    \item steps must appear in dependency order without forward references;
    \item each step value must be consistent with its operation and arguments;
          and
    \item the final step value must equal the reported answer.
\end{itemize}

These requirements constrain the semantic structure of the computation graph,
while the carrier-specific contract determines only how that structure is
serialized.

\subsubsection{Same Artifact Across Seven Representations}
\label{app:representation-examples}

Table~\ref{tab:representation-examples} illustrates the controlled-comparison
design using the validated TabMWP instance
\texttt{tabmwp-default-train-28054}. Its Canonical IR contains the answer
$3521$, the semantic input
\texttt{caribbean\_price}$=503$, and the computation
\texttt{R1} $=$
\texttt{multiply}(\texttt{caribbean\_price}, $7$) $=3521$.
The semantic artifact is fixed across all rows; only the carrier syntax changes.

\begin{table}[H]
\caption{The same Canonical IR rendered in the seven SERIAL-7 representation
families. Ellipses abbreviate carrier-specific graph details while preserving
the same underlying semantic content.}
\centering
\small
\renewcommand{\arraystretch}{1.25}
\begin{tabular}{
@{}
>{\raggedright\arraybackslash}p{0.14\linewidth}
>{\raggedright\arraybackslash}p{0.78\linewidth}
@{}
}
\toprule
\textbf{Format} &
\textbf{Representative realization} \\
\midrule

JSON &
\texttt{\{"answer": 3521, "inputs": [...], "steps": [...]\}} \\

XML &
\texttt{<response><answer>3521</answer> ... </response>} \\

Markdown &
\texttt{\# Result; - Answer: 3521; \#\# Inputs; \#\# Steps} \\

HTML &
\texttt{<main><output id="answer">3521</output> ... </main>} \\

LaTeX &
\texttt{\textbackslash begin\{response\}
\textbackslash Answer\{3521\} ...
\textbackslash end\{response\}} \\

YAML &
\texttt{answer: 3521; inputs: ...; steps: ...} \\

TOML &
\texttt{answer = 3521; [[inputs]]; [[steps]]} \\

\bottomrule
\end{tabular}
\label{tab:representation-examples}
\end{table}

\subsubsection{Example of Composed Atomic Constraints}
\label{app:feature-example}

As an example, consider the same JSON artifact above under an L2 configuration
containing \texttt{answer\_object} and
\texttt{numeric\_as\_string}. The semantic artifact is unchanged, but the
representation becomes:

\begin{verbatim}
{
  "answer": {
    "value": "3521"
  },
  "inputs": [
    {
      "name": "caribbean_price",
      "value": "503"
    }
  ],
  "steps": [
    {
      "id": "R1",
      "op": "multiply",
      "args": ["caribbean_price", "7"],
      "value": "3521"
    }
  ]
}
\end{verbatim}

Here, \texttt{answer\_object} changes the shape of the answer slot and
\texttt{numeric\_as\_string} changes the surface type of numeric values.
Neither rule changes the underlying answer, semantic inputs, or computation.

\subsection{Preserve: Representation Conversion}
\label{app:preserve-prompts}

\textsc{Preserve} removes semantic derivation by providing the model with an
already instantiated source document. Unlike \textsc{Realize}, it uses the
base target representation contract without activating sampled atomic
representation constraints. The model rewrites the source artifact in the
target representation while preserving its required content.

\PromptTitle{Preserve: Assembled User Message}
\begin{PromptBlock}
## Input Document
{source document rendered in the source format}

## Instructions
Rewrite the input document above in the {target format} while preserving exactly the same content.

Do not change or recompute any value.
Preserve all names and ordering.

{base target representation contract}

## Constraints
Output only the requested structure and fields; add no other text or fields.
Before output, verify that all values match the input document.
\end{PromptBlock}

\textsc{Preserve} reuses the same representation-specific parser and base-contract
checks as \textsc{Realize}, but sampled atomic-rule checks are not activated.
Because the source artifact is already instantiated, the model is not required
to derive the answer; evaluation instead verifies faithful source--target
preservation under the base target representation contract.
\section{Executable Verification Details}
\label{app:executable-verification}

\begin{figure*}[t]
    \centering
    \includegraphics[width=\textwidth]{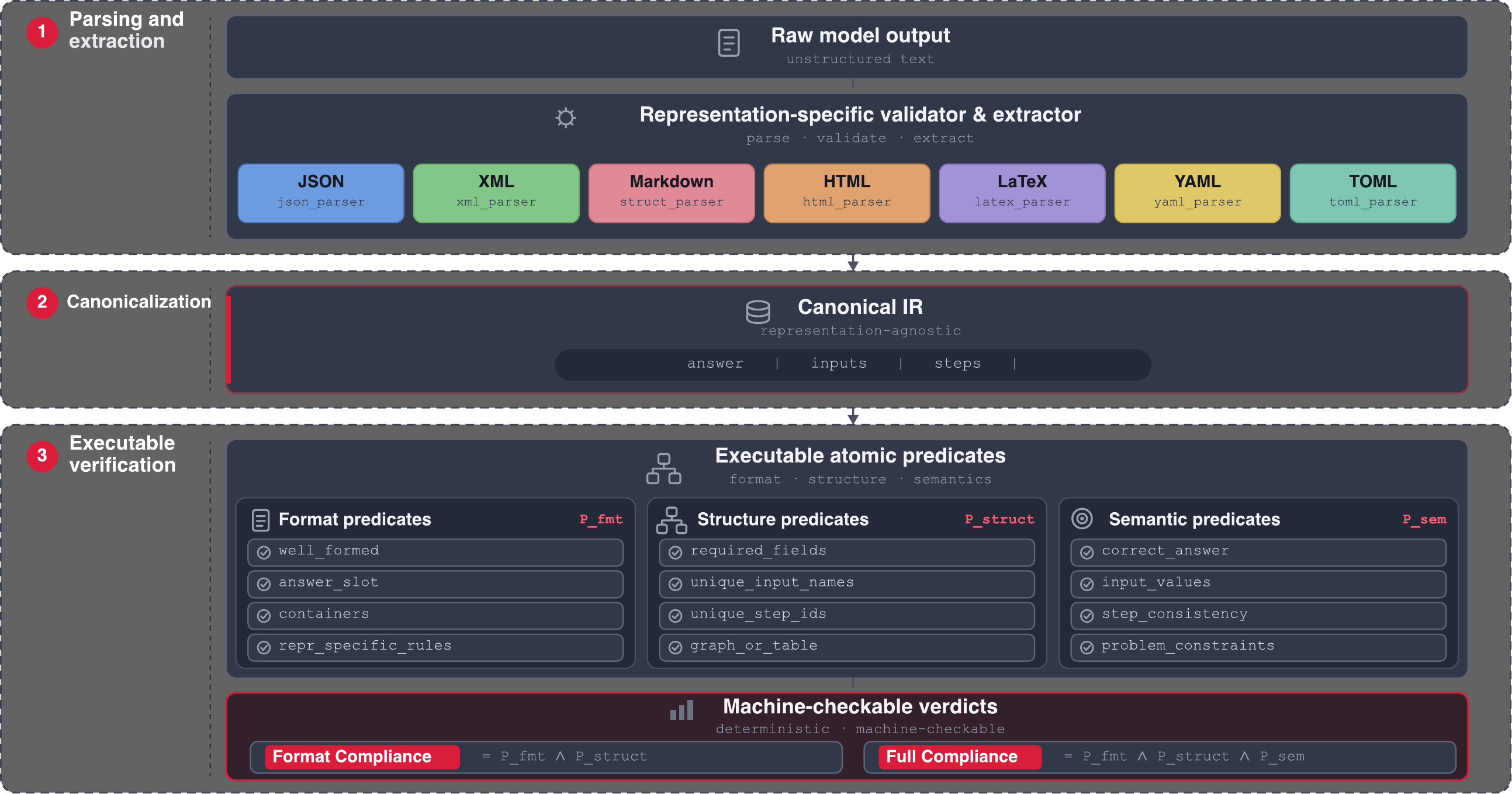}
    \caption{
    \textsc{Serial-7} executable verification pipeline for \textsc{Realize} and
    \textsc{Preserve}. Representation-specific validators and extractors map
    model outputs to a shared Canonical IR, after which executable predicates
    produce machine-checkable compliance verdicts.
    }
    \label{fig:serial7-eval-pipeline}
\end{figure*}

SERIAL-7 evaluates structured outputs through the executable pipeline in
Figure~\ref{fig:serial7-eval-pipeline}. Representation-specific front ends
validate and extract model outputs into a shared Canonical IR, after which
shared predicates evaluate format, structure, and semantics.

The verifier separates four namespaces:
\texttt{format.*}, \texttt{format.feature.*}, \texttt{structure.*}, and
\texttt{semantic.*}. The first two jointly define representation compliance:
\texttt{format.*} contains the active base-carrier requirements, while
\texttt{format.feature.*} contains one predicate for each active atomic
constraint. When an atomic constraint intentionally changes part of the base
carrier contract, the incompatible base expectation is replaced by the
corresponding rule-aware expectation rather than enforced simultaneously.
\textsc{Derive} activates no mechanical format or structure checks and is
evaluated only for final-answer correctness by the semantic judge in
Appendix~\ref{app:derive-prompt}.

For visual simplicity, Figure~\ref{fig:serial7-eval-pipeline} shows
Format Compliance and Full Compliance as verifier aggregates. HSR is derived
from Format Compliance and the final-answer correctness predicate, as formally
defined in Appendix~\ref{app:normalization-scoring}.

\subsection{Atomic Check Inventory}
\label{app:atomic-check-inventory}

The required format, feature, and structure checks are determined from the
target representation, artifact type, and active atomic-rule set before the
model output is scored. Thus, a parse failure cannot improve Format Compliance
by shrinking the set of required representation checks. Semantic
graph-consistency is evaluated for graph-bearing artifacts when an executable
graph can be recovered; this semantic diagnostic does not affect Format
Compliance or HSR.

\begin{table}[H]
\caption{Atomic check namespaces used by the executable verifier.}
\label{tab:atomic-check-inventory}
\centering
\small
\renewcommand{\arraystretch}{1.08}
\setlength{\tabcolsep}{4pt}

\begin{tabular}{
@{}
>{\raggedright\arraybackslash}p{0.21\linewidth}
>{\raggedright\arraybackslash}p{0.32\linewidth}
>{\raggedright\arraybackslash}p{0.39\linewidth}
@{}
}
\toprule
\textbf{Namespace} &
\textbf{Atomic checks} &
\textbf{Purpose} \\
\midrule

\texttt{format.*} &
\texttt{clean\_boundary},
\texttt{well\_formed},
\texttt{native\_syntax},
\texttt{root\_container},
\texttt{answer\_slot},
\texttt{inputs\_container},
\texttt{steps\_container},
\texttt{escaping} &
Validate document boundaries, parseability, required slots and containers,
native syntax, and escaping under the active representation contract. \\[2pt]

\texttt{format.feature.*} &
One check per active atomic rule &
Validate the sampled atomic representation rule(s) used in
\textsc{Realize}. \\[2pt]

\texttt{structure.*} &
\texttt{unique\_input\_names},
\texttt{unique\_step\_ids},
\texttt{input\_fields\_present},
\texttt{step\_fields\_present} &
Check completeness and uniqueness of extracted input and step records. \\[2pt]

\texttt{semantic.*} &
\texttt{answer},
\texttt{input\_values},
\texttt{step\_consistent} &
Check final-answer correctness, required input values, and executable graph
consistency when applicable. \\

\bottomrule
\end{tabular}
\end{table}

Table~\ref{tab:format-check-denominators} reports the nominal base
\texttt{format.*} inventory when no atomic rule modifies the corresponding
carrier requirement. When an active rule changes a base representation
requirement, the incompatible base predicate is replaced by its rule-aware
counterpart before scoring.

\begin{table}[H]
\caption{
Nominal active \texttt{format.*} checks before rule-aware substitution.
}
\label{tab:format-check-denominators}
\centering
\begin{tabular}{@{}lrrrrrrr@{}}
\toprule
& JSON & XML & Markdown & HTML & \LaTeX & YAML & TOML \\
\midrule
Answer-only & 4 & 5 & 5 & 6 & 5 & 5 & 4 \\
Graph-bearing & 6 & 7 & 7 & 8 & 7 & 7 & 6 \\
\bottomrule
\end{tabular}
\end{table}

A level-$\ell$ \textsc{Realize} item activates exactly $\ell$ feature
predicates in addition to its rule-aware base contract. Structure checks are
defined over records recovered from the model output, so alternative valid
decompositions are not penalized solely for differing in length from the gold
graph. Unit metadata is construction-only and is excluded from the scored
Canonical IR.

\subsection{Representation-Specific Parsers}
\label{app:representation-parsers}

SERIAL-7 does not use one generic parser for all outputs. Each representation
family has its own executable front end for native parsing and extraction.
When no atomic rule is active, the base carrier contract is used directly.
When one or more atomic rules are active, the expected carrier contract is
modified before validation so that a requested representation transformation
is not simultaneously penalized for violating the superseded base form.

Algorithm~\ref{alg:carrier-analysis} expands the high-level evaluation
procedure in Section~\ref{sec:evaluation_protocol}. The helper procedures
\Call{CarrierAnalyze} and \Call{FeatureAnalyze} correspond to the two branches
shown in the main-text pseudocode, while \Call{AtomicCheck} evaluates each
active atomic rule independently.

\begin{breakablealgorithm}
\caption{Carrier-Specific Mechanical Verification and Extraction}
\label{alg:carrier-analysis}

\begin{algorithmic}[1]

\Require target format $f$, model output $\hat{y}$,
gold Canonical IR $z$, active atomic rules $\mathcal{A}$
\Ensure Canonical IR $\hat{z}$, base verdicts $\mathbf{b}$,
feature verdicts $\mathbf{q}$

\State $g
    \gets
    (z.\mathtt{inputs}\neq\emptyset)
    \land
    (z.\mathtt{steps}\neq\emptyset)$
\State $\nu
    \gets
    \Call{IsNumericAnswer}{z.\mathtt{answer}}$

\State $\mathcal{B}
    \gets
    \Call{BaseChecks}{f,g,\mathcal{A}}$
\State $\mathbf{b}
    \gets
    \Call{InitializeFalse}{\mathcal{B}}$

\If{$\mathcal{A}\neq\emptyset$}
    \State $(\hat{z},\mathbf{b})
        \gets
        \Call{FeatureAnalyze}
        {f,\hat{y},g,\nu,\mathcal{A},\mathbf{b}}$
\Else
    \State $(\hat{z},\mathbf{b})
        \gets
        \Call{CarrierAnalyze}
        {f,\hat{y},g,\nu,\mathcal{B},\mathbf{b}}$
\EndIf

\State $\mathbf{b}[\mathrm{clean\_boundary}]
    \gets
    \Call{CleanBoundary}{f,\hat{y},\mathcal{A}}$

\State $\mathbf{q}\gets\emptyset$
\ForAll{$a\in\mathcal{A}$}
    \State $\mathbf{q}[a]
        \gets
        \Call{AtomicCheck}{f,a,\hat{y},\hat{z}}$
\EndFor

\State \Return $(\hat{z},\mathbf{b},\mathbf{q})$

\Statex

\Procedure{CarrierAnalyze}
{$f,\hat{y},g,\nu,\mathcal{B},\mathbf{b}$}

    \If{$f=\mathtt{json}$}
        \State $x\gets\Call{StrictJSONParse}{\hat{y}}$
        \State $\mathbf{b}[\mathrm{well\_formed}]
            \gets(x\neq\bot)$
        \If{$x\neq\bot$}
            \State $\mathbf{b}[\mathrm{root\_container}]
                \gets\Call{IsMapping}{x}$
            \State $\mathbf{b}[\mathrm{answer\_slot}]
                \gets\Call{HasTypedAnswer}{x,\nu}$
            \If{$g$}
                \State $\mathbf{b}[\mathrm{inputs\_container}]
                    \gets
                    \Call{IsSequenceField}{x,\mathtt{inputs}}$
                \State $\mathbf{b}[\mathrm{steps\_container}]
                    \gets
                    \Call{IsSequenceField}{x,\mathtt{steps}}$
            \EndIf
        \EndIf

    \ElsIf{$f=\mathtt{xml}$}
        \State $x\gets\Call{SafeXMLParse}{\hat{y}}$
        \State $\mathbf{b}[\mathrm{well\_formed}]
            \gets(x\neq\bot)$
        \If{$x\neq\bot$}
            \State $\mathbf{b}[\mathrm{root\_container}]
                \gets\Call{SingleResponseRoot}{x}$
            \State $\mathbf{b}[\mathrm{answer\_slot}]
                \gets\Call{SingleDirectAnswer}{x}$
            \If{$g$}
                \State $\mathbf{b}[\mathrm{inputs\_container}]
                    \gets
                    \Call{SingleDirectContainer}
                    {x,\mathtt{inputs}}$
                \State $\mathbf{b}[\mathrm{steps\_container}]
                    \gets
                    \Call{SingleDirectContainer}
                    {x,\mathtt{steps}}$
            \EndIf
        \EndIf

    \ElsIf{$f=\mathtt{markdown}$}
        \State $x\gets\Call{MarkdownStateMachine}{\hat{y}}$
        \State $\mathbf{b}[\mathrm{well\_formed}]
            \gets\Call{NoStrayLines}{x}$
        \State $\mathbf{b}[\mathrm{root\_container}]
            \gets\Call{HasResultHeading}{x}$
        \State $\mathbf{b}[\mathrm{answer\_slot}]
            \gets\Call{HasAnswerField}{x}$
        \If{$g$}
            \State $\mathbf{b}[\mathrm{inputs\_container}]
                \gets
                \Call{SingleSection}{x,\mathtt{inputs}}$
            \State $\mathbf{b}[\mathrm{steps\_container}]
                \gets
                \Call{SingleSection}{x,\mathtt{steps}}$
        \EndIf

    \ElsIf{$f=\mathtt{html}$}
        \State $x\gets\Call{StrictLIFOHTMLParse}{\hat{y}}$
        \State $\mathbf{b}[\mathrm{well\_formed}]
            \gets\Call{TagsBalanced}{x}$
        \State $\mathbf{b}[\mathrm{root\_container}]
            \gets\Call{SingleMain}{x}$
        \State $\mathbf{b}[\mathrm{answer\_slot}]
            \gets\Call{AnswerNestedInMain}{x}$
        \If{$g$}
            \State $\mathbf{b}[\mathrm{inputs\_container}]
                \gets\Call{InputsNestedInMain}{x}$
            \State $\mathbf{b}[\mathrm{steps\_container}]
                \gets\Call{StepsNestedInMain}{x}$
        \EndIf

    \ElsIf{$f=\mathtt{latex}$}
        \State $x
            \gets
            \Call{ScanMacrosAndEnvironments}{\hat{y}}$
        \State $\mathbf{b}[\mathrm{well\_formed}]
            \gets\Call{BalancedEnvironments}{x}$
        \State $\mathbf{b}[\mathrm{root\_container}]
            \gets\Call{SingleResponseEnvironment}{x}$
        \State $\mathbf{b}[\mathrm{answer\_slot}]
            \gets\Call{SingleAnswerMacro}{x}$
        \If{$g$}
            \State $\mathbf{b}[\mathrm{inputs\_container}]
                \gets\Call{SingleInputsEnvironment}{x}$
            \State $\mathbf{b}[\mathrm{steps\_container}]
                \gets\Call{SingleStepsEnvironment}{x}$
        \EndIf

    \ElsIf{$f=\mathtt{yaml}$}
        \State $x\gets\Call{RestrictedYAMLParse}{\hat{y}}$
        \State $\mathbf{b}[\mathrm{well\_formed}]
            \gets
            (x\neq\bot)
            \land
            \Call{NoDocumentMarker}{\hat{y}}$
        \If{$x\neq\bot$}
            \State $\mathbf{b}[\mathrm{root\_container}]
                \gets\Call{IsMapping}{x}$
            \State $\mathbf{b}[\mathrm{answer\_slot}]
                \gets
                \Call{HasTypedNativeAnswer}{x,\nu}$
            \If{$g$}
                \State $\mathbf{b}[\mathrm{inputs\_container}]
                    \gets
                    \Call{IsBlockSequenceField}
                    {x,\mathtt{inputs}}$
                \State $\mathbf{b}[\mathrm{steps\_container}]
                    \gets
                    \Call{IsBlockSequenceField}
                    {x,\mathtt{steps}}$
            \EndIf
        \EndIf

    \ElsIf{$f=\mathtt{toml}$}
        \State $x\gets\Call{RestrictedTOMLParse}{\hat{y}}$
        \State $\mathbf{b}[\mathrm{well\_formed}]
            \gets(x\neq\bot)$
        \If{$x\neq\bot$}
            \State $\mathbf{b}[\mathrm{root\_container}]
                \gets\Call{IsMapping}{x}$
            \State $\mathbf{b}[\mathrm{answer\_slot}]
                \gets
                \Call{HasTypedTopLevelAnswer}{x,\nu}$
            \If{$g$}
                \State $\mathbf{b}[\mathrm{inputs\_container}]
                    \gets
                    \Call{HasInputArrayTables}{x}$
                \State $\mathbf{b}[\mathrm{steps\_container}]
                    \gets
                    \Call{HasStepArrayTables}{x}$
            \EndIf
        \EndIf
    \EndIf

    \If{$\mathtt{native\_syntax}\in\mathcal{B}$}
        \State $\mathbf{b}[\mathrm{native\_syntax}]
            \gets
            \Call{HasNativeSyntax}{f,\hat{y}}$
    \EndIf

    \If{$\mathtt{escaping}\in\mathcal{B}$}
        \State $\mathbf{b}[\mathrm{escaping}]
            \gets
            \Call{HasValidEscaping}{f,\hat{y}}$
    \EndIf

    \State $\hat{z}
        \gets
        \Call{BaseExtract}{f,\hat{y}}$

    \State \Return $(\hat{z},\mathbf{b})$

\EndProcedure

\Statex

\Procedure{FeatureAnalyze}
{$f,\hat{y},g,\nu,\mathcal{A},\mathbf{b}$}

    \State $x
        \gets
        \Call{NativeParse}{f,\hat{y}}$

    \State $\mathbf{b}
        \gets
        \Call{EvaluateActiveBaseChecks}
        {f,x,\hat{y},g,\nu,\mathcal{A},\mathbf{b}}$

    \State $\hat{z}
        \gets
        \Call{FeatureAwareExtract}
        {f,\hat{y},\mathcal{A}}$

    \State \Return $(\hat{z},\mathbf{b})$

\EndProcedure

\Statex

\Procedure{AtomicCheck}{$f,a,\hat{y},\hat{z}$}

    \State $\rho\gets\mathrm{false}$

    \If{$f=\mathtt{json}$}
        \If{$a=\mathtt{inputs\_as\_map}$}
            \State $\rho
                \gets\Call{NameKeyedValueMap}{\hat{y}}$
        \ElsIf{$a=\mathtt{steps\_as\_map}$}
            \State $\rho
                \gets\Call{IdKeyedStepMap}{\hat{y}}$
        \ElsIf{$a=\mathtt{args\_as\_objects}$}
            \State $\rho
                \gets\Call{AllArgsAreRefObjects}{\hat{y}}$
        \ElsIf{$a=\mathtt{answer\_object}$}
            \State $\rho
                \gets\Call{AnswerIsValueObject}{\hat{y}}$
        \ElsIf{$a=\mathtt{numeric\_as\_string}$}
            \State $\rho
                \gets\Call{AllNumbersQuoted}{\hat{y}}$
        \ElsIf{$a=\mathtt{key\_order}$}
            \State $\rho
                \gets\Call{RecordKeysAlphabetical}{\hat{y}}$
        \EndIf

    \ElsIf{$f=\mathtt{xml}$}
        \If{$a=\mathtt{args\_as\_attribute}$}
            \State $\rho
                \gets
                \Call{EveryStepHasArgsAttribute}{\hat{y}}
                \land
                \neg\Call{HasArgChild}{\hat{y}}$
        \ElsIf{$a=\mathtt{namespace}$}
            \State $\rho
                \gets\Call{HasRequiredNamespace}{\hat{y}}$
        \ElsIf{$a=\mathtt{sibling\_order}$}
            \State $\rho
                \gets\Call{StepsBeforeInputs}{\hat{y}}$
        \ElsIf{$a=\mathtt{value\_as\_attribute}$}
            \State $\rho
                \gets\Call{InputValuesUseAttributes}{\hat{y}}$
        \ElsIf{$a=\mathtt{answer\_last}$}
            \State $\rho
                \gets\Call{AnswerAfterGraph}{\hat{y}}$
        \EndIf

    \ElsIf{$f=\mathtt{markdown}$}
        \If{$a=\mathtt{inputs\_as\_table}$}
            \State $\rho
                \gets
                \Call{PipeTableWithSeparator}
                {\hat{y},\mathtt{inputs}}$
        \ElsIf{$a=\mathtt{steps\_as\_table}$}
            \State $\rho
                \gets
                \Call{PipeTableWithSeparator}
                {\hat{y},\mathtt{steps}}$
        \ElsIf{$a=\mathtt{fence}$}
            \State $\rho
                \gets\Call{StepsInTypedFence}{\hat{y}}$
        \ElsIf{$a=\mathtt{inline\_code\_ids}$}
            \State $\rho
                \gets
                \Call{AllIdentifiersBackticked}
                {\hat{y},\hat{z}}$
        \ElsIf{$a=\mathtt{heading\_depth}$}
            \State $\rho
                \gets\Call{RequiredHeadingDepth}{\hat{y}}$
        \ElsIf{$a=\mathtt{blockquote\_answer}$}
            \State $\rho
                \gets\Call{AnswerIsBlockquote}{\hat{y}}$
        \ElsIf{$a=\mathtt{bold\_names}$}
            \State $\rho
                \gets
                \Call{AllInputNamesBold}{\hat{y},\hat{z}}$
        \EndIf

    \ElsIf{$f=\mathtt{html}$}
        \If{$a=\mathtt{table\_for\_inputs}$}
            \State $\rho
                \gets
                \Call{TableWithHeader}
                {\hat{y},\mathtt{inputs}}$
        \ElsIf{$a=\mathtt{table\_for\_steps}$}
            \State $\rho
                \gets
                \Call{TableWithHeader}
                {\hat{y},\mathtt{steps}}$
        \ElsIf{$a=\mathtt{dl\_for\_answer}$}
            \State $\rho
                \gets\Call{AnswerInDefinitionList}{\hat{y}}$
        \ElsIf{$a=\mathtt{semantic\_element}$}
            \State $\rho
                \gets\Call{FigureWithCaption}{\hat{y}}$
        \ElsIf{$a=\mathtt{strong\_answer}$}
            \State $\rho
                \gets\Call{AnswerInsideStrong}{\hat{y}}$
        \ElsIf{$a=\mathtt{heading\_hierarchy}$}
            \State $\rho
                \gets\Call{ValidHeadingHierarchy}{\hat{y}}$
        \EndIf

    \ElsIf{$f=\mathtt{latex}$}
        \If{$a=\mathtt{tabular\_inputs}$}
            \State $\rho
                \gets
                \Call{HasTwoColumnInputTable}{\hat{y},\hat{z}}
                \land
                \Call{HasHorizontalRule}{\hat{y}}$
        \ElsIf{$a=\mathtt{enumerate\_steps}$}
            \State $\rho
                \gets\Call{StepsUseEnumerate}{\hat{y}}$
        \ElsIf{$a=\mathtt{siunitx}$}
            \State $\rho
                \gets
                \Call{AllNumbersWrappedByNum}{\hat{y}}
                \land
                \neg\Call{HasBareNumber}{\hat{y}}$
        \ElsIf{$a=\mathtt{math\_mode}$}
            \State $\rho
                \gets\Call{ValuesUseMathMode}{\hat{y}}$
        \ElsIf{$a=\mathtt{texttt\_ids}$}
            \State $\rho
                \gets
                \Call{AllIdentifiersInTexttt}
                {\hat{y},\hat{z}}$
        \ElsIf{$a=\mathtt{bold\_ids}$}
            \State $\rho
                \gets
                \Call{AllStepIdsBold}{\hat{y},\hat{z}}$
        \EndIf

    \ElsIf{$f=\mathtt{yaml}$}
        \If{$a=\mathtt{inputs\_as\_map}$}
            \State $\rho
                \gets\Call{NameKeyedValueMap}{\hat{y}}$
        \ElsIf{$a=\mathtt{steps\_as\_map}$}
            \State $\rho
                \gets\Call{IdKeyedStepMap}{\hat{y}}$
        \ElsIf{$a=\mathtt{answer\_object}$}
            \State $\rho
                \gets\Call{AnswerIsValueMap}{\hat{y}}$
        \ElsIf{$a=\mathtt{numeric\_as\_string}$}
            \State $\rho
                \gets\Call{AllNumbersDoubleQuoted}{\hat{y}}$
        \ElsIf{$a=\mathtt{key\_order}$}
            \State $\rho
                \gets\Call{MappingKeysAlphabetical}{\hat{y}}$
        \ElsIf{$a=\mathtt{args\_as\_objects}$}
            \State $\rho
                \gets\Call{AllArgsAreRefMaps}{\hat{y}}$
        \ElsIf{$a=\mathtt{flow\_args}$}
            \State $\rho
                \gets
                \Call{ArgsUseFlowStyle}{\hat{y}}
                \land
                \neg\Call{ArgsUseBlockStyle}{\hat{y}}$
        \EndIf

    \ElsIf{$f=\mathtt{toml}$}
        \If{$a=\mathtt{inputs\_as\_table}$}
            \State $\rho
                \gets
                \Call{NameKeyedInputTables}{\hat{y}}$
        \ElsIf{$a=\mathtt{steps\_as\_table}$}
            \State $\rho
                \gets
                \Call{IdKeyedStepTables}{\hat{y}}$
        \ElsIf{$a=\mathtt{answer\_table}$}
            \State $\rho
                \gets
                \Call{AnswerStoredInTable}{\hat{y}}$
        \ElsIf{$a=\mathtt{numeric\_as\_string}$}
            \State $\rho
                \gets
                \Call{AllNumbersDoubleQuoted}{\hat{y}}$
        \ElsIf{$a=\mathtt{key\_order}$}
            \State $\rho
                \gets
                \Call{KeysAlphabeticalWithinTables}{\hat{y}}$
        \ElsIf{$a=\mathtt{args\_as\_tables}$}
            \State $\rho
                \gets
                \Call{ArgsAreInlineRefTables}{\hat{y}}$
        \EndIf
    \EndIf

    \State \Return $\rho$

\EndProcedure

\end{algorithmic}
\end{breakablealgorithm}
The base and feature-aware branches use the same native representation family
but differ in the expected contract applied during validation. In particular,
an atomic rule that changes a base slot, container, or scalar representation
supersedes the incompatible base expectation for that component.
The corresponding atomic predicate is still evaluated separately, so the
feature requirement remains directly observable in the verifier output.

Feature-aware extraction may recover semantic content even when a local
representation predicate fails. This separation prevents a local surface
violation from being automatically counted as an unrelated downstream
semantic error.

\subsection{Normalization and Scoring Rules}
\label{app:normalization-scoring}

After extraction, every structured output is projected to the scored Canonical
IR, $\{\texttt{answer},\texttt{inputs},\texttt{steps}\}$. The interpretation
of this IR is task-specific: \textsc{Realize} evaluates whether the generated
artifact contains the correct answer under the requested representation
contract, whereas \textsc{Preserve} requires preservation of the instantiated
source artifact under the benchmark-defined invariants. These invariants include
the answer, ordered input identifiers and values, and, when present, ordered
step identifiers, operators, argument references, and values. Consequently,
identifier renaming or otherwise harmless reordering fails the preservation
predicate even when semantic equivalence is retained.

\begin{table}[H]
\caption{Representation-independent normalization and scoring rules.}
\label{tab:normalization-rules}
\centering
\begin{tabular}{
@{}
>{\raggedright\arraybackslash}p{0.22\linewidth}
>{\raggedright\arraybackslash}p{0.69\linewidth}
@{}
}
\toprule
\textbf{Component} &
\textbf{Normalization / scoring rule} \\
\midrule

Numeric values &
Integer gold values require exact equality; non-integers use relative
tolerance $10^{-6}$ and absolute tolerance $10^{-9}$ unless overridden.
Representation-type errors are charged to \texttt{format.*}. \\

Text values &
Case-fold, trim and collapse whitespace, and remove trailing sentence
punctuation where the corresponding semantic field is evaluated by normalized
text equality. Identifiers and references used by the preservation predicate
remain part of the Canonical IR and are compared source-to-target. \\

\textsc{Realize} input semantics &
Extracted inputs are matched to required gold inputs by value rather than by
arbitrary gold identifier. Exact reproduction of gold input names is therefore
not required for \textsc{Realize} semantic correctness. \\

\textsc{Realize} graph consistency &
When a graph-bearing output can be recovered, execute the model's own graph
rather than requiring step-by-step equality with the gold decomposition.
References must resolve, step values must execute correctly, and the final
step must agree with the model's answer. This predicate is a diagnostic
semantic check and is not part of \textsc{Realize} HSR. \\

\textsc{Preserve} input preservation &
The ordered sequence of input identifiers and corresponding values extracted
from the target must equal that of the instantiated source artifact. A target
that preserves only the values while changing, omitting, or reordering the
required identifiers does not satisfy the preservation predicate. \\

\textsc{Preserve} step preservation &
For graph-bearing artifacts, the ordered source and target step records must
agree in their step identifiers, operators, argument references, and values.
Missing, altered, or reordered required step content causes the preservation
predicate to fail. \\

Unrecoverable optional graph diagnostic &
For \textsc{Realize}, if an emitted graph cannot be recovered reliably enough
for execution, the graph-consistency diagnostic is marked non-evaluable rather
than introducing an additional HSR failure. This rule does not weaken
\textsc{Preserve}: when the source artifact contains required step content,
failure to recover or preserve that content causes the preservation predicate
to fail. \\

\textsc{Derive} &
No mechanical representation score is defined; final-answer correctness is
evaluated by the semantic judge. \\

\bottomrule
\end{tabular}
\end{table}

Let
$\mathcal{C}_{\mathrm{base}}$ denote the active rule-aware predicates from
\texttt{format.*},
$\mathcal{C}_{\mathrm{feat}}$ the active predicates from
\texttt{format.feature.*}, and
$\mathcal{C}_{\mathrm{struct}}$ the active predicates from
\texttt{structure.*}.
Format Compliance requires all active representation and structural
predicates to pass:
\[
\mathrm{FC}
=
\bigwedge_{p\in
\mathcal{C}_{\mathrm{base}}
\cup
\mathcal{C}_{\mathrm{feat}}
\cup
\mathcal{C}_{\mathrm{struct}}}
p.
\]

HSR combines Format Compliance with a task-specific semantic success
predicate. For a structured task $\tau$, we write
\[
\mathrm{HSR}_{\tau}
=
\mathrm{FC}
\land
P_{\tau}.
\]

For \textsc{Realize}, the primary semantic requirement is final-answer
correctness:
\[
P_{\mathrm{Realize}}
=
p_{\mathrm{answer}},
\]
and therefore
\[
\mathrm{HSR}_{\mathrm{Realize}}
=
\mathrm{FC}
\land
p_{\mathrm{answer}}.
\]

For \textsc{Preserve}, final-answer correctness alone is insufficient.
Let
$p_{\mathrm{inputs}}^{\mathrm{pres}}$
denote source--target equality of the ordered input identifiers and values,
and let
$p_{\mathrm{steps}}^{\mathrm{pres}}$
denote source--target equality of the ordered step records, including step
identifiers, operators, argument references, and values when steps are
present. We define
\[
P_{\mathrm{pres}}
=
p_{\mathrm{answer}}
\land
p_{\mathrm{inputs}}^{\mathrm{pres}}
\land
p_{\mathrm{steps}}^{\mathrm{pres}},
\]
where
$p_{\mathrm{steps}}^{\mathrm{pres}}$ is vacuously true for artifacts without
step records. Accordingly,
\[
P_{\mathrm{Preserve}}
=
P_{\mathrm{pres}},
\]
and
\[
\mathrm{HSR}_{\mathrm{Preserve}}
=
\mathrm{FC}
\land
P_{\mathrm{pres}}.
\]

Thus, a \textsc{Preserve} output does not count as a holistic success merely
because it is well-formed and retains the correct final answer. It must also
preserve the instantiated source artifact required by the Canonical IR.

For diagnostic reporting, let
$\mathcal{C}_{\mathrm{sem}}^{\tau}$ denote all active evaluable semantic
predicates for task $\tau$. Full Compliance requires every active
representation, structural, and evaluable semantic predicate:
\[
\mathrm{Full}_{\tau}
=
\bigwedge_{p\in
\mathcal{C}_{\mathrm{base}}
\cup
\mathcal{C}_{\mathrm{feat}}
\cup
\mathcal{C}_{\mathrm{struct}}
\cup
\mathcal{C}_{\mathrm{sem}}^{\tau}}
p.
\]

For \textsc{Realize},
$\mathcal{C}_{\mathrm{sem}}^{\mathrm{Realize}}$
includes final-answer correctness and any additional evaluable semantic
diagnostics, such as required input-value correctness and executable
graph consistency. For \textsc{Preserve},
$\mathcal{C}_{\mathrm{sem}}^{\mathrm{Preserve}}$
includes the preservation predicates above together with any additional
evaluable diagnostics.

Consequently, Format Compliance isolates adherence to the active
representation contract and required structure. HSR adds the primary
task-specific semantic requirement:
final-answer correctness for \textsc{Realize}, and full source--target
artifact preservation for \textsc{Preserve}. Full Compliance is a stricter
diagnostic aggregate over all other active evaluable semantic predicates.

\end{document}